%% file: main.tex
\documentclass{article}

\usepackage{microtype}
\usepackage{graphicx}
\usepackage{subfigure}
\usepackage{booktabs} 
\usepackage{bbm} 
\usepackage{amsmath}

\usepackage{hyperref}

\newcommand{\xxnote}[3]{}
\ifx\hidenotes\undefined
  \renewcommand{\xxnote}[3]{\color{#2}{#1: #3}}
\fi

\usepackage[accepted]{icml2020}
\input{math_commands.tex}

\icmltitlerunning{Hierarchically Decoupled Imitation for Morphological Transfer}

\begin{document}

\twocolumn[
\icmltitle{Hierarchically Decoupled Imitation for Morphological Transfer}

\icmlsetsymbol{equal}{*}

\begin{icmlauthorlist}
\icmlauthor{Donald J. Hejna III}{cal}
\icmlauthor{Pieter Abbeel}{cal}
\icmlauthor{Lerrel Pinto}{cal,nyu}
\end{icmlauthorlist}

\icmlaffiliation{cal}{Department of EECS, University of California, Berkeley}
\icmlaffiliation{nyu}{Computer Science, New York University}
\icmlcorrespondingauthor{Donald J. Hejna III}{jhejna@berkeley.edu}
\icmlkeywords{Morphology Transfer, Reinforcement Learning, Machine Learning}
\vskip 0.3in
]
\printAffiliationsAndNotice{}  

\input{abstract.tex}
\input{introduction.tex}

\input{method.tex}
\input{results.tex}
\input{related_work.tex}

\input{conclusions.tex}


\bibliography{references}
\bibliographystyle{icml2020}
\newpage
\onecolumn
\input{appendix.tex}
\end{document}

%% file: math_commands.tex
\usepackage{amsmath,amsfonts,bm}

\def\Figref#1{Figure~\ref{#1}}

\def\Secref#1{Section~\ref{#1}}

\def\eqref#1{equation~\ref{#1}}

\def\Algref#1{Algorithm~\ref{#1}}

\def\1{\bm{1}}

\DeclareMathAlphabet{\mathsfit}{\encodingdefault}{\sfdefault}{m}{sl}
\SetMathAlphabet{\mathsfit}{bold}{\encodingdefault}{\sfdefault}{bx}{n}

\def\sG{{\mathbb{G}}}

\newcommand{\E}{\mathbb{E}}

\DeclareMathOperator*{\expect}{\E}

\newcommand{\Tau}{\mathcal{T}}

%% file: abstract.tex
\begin{abstract}

Learning long-range behaviors on complex high-dimensional agents is a fundamental problem in robot learning. For such tasks, we argue that transferring learned information from a morphologically simpler agent can massively improve the sample efficiency of a more complex one. To this end, we propose a hierarchical decoupling of policies into two parts: an independently learned low-level policy and a transferable high-level policy. To remedy poor transfer performance due to mismatch in morphologies, we contribute two key ideas. First, we show that incentivizing a complex agent's low-level to imitate a simpler agent's low-level significantly improves zero-shot high-level transfer. Second, we show that KL-regularized training of the high level stabilizes learning and prevents mode-collapse. Finally, on a suite of publicly released navigation and manipulation environments, we demonstrate the applicability of hierarchical transfer on long-range tasks across morphologies. Our code and videos can be found at \url{https://sites.google.com/berkeley.edu/morphology-transfer}.

\end{abstract}

%% file: introduction.tex
\section{Introduction}
How should one use Reinforcement Learning (RL) to train their new four-legged walking robot? Training a robot from scratch is often infeasible as current RL algorithms typically require millions~\cite{schulman2017proximal,haarnoja2018soft} to billions~\cite{baker2019emergent} of samples. Moreover, robots are expensive and slow, which further limits the applicability of learning from scratch. Because of this, tackling the challenge of high sample complexity has received significant interest and prompted a wide array of solutions ranging from off-policy learning~\cite{lillicrap2015continuous} to temporal abstractions in learning~\cite{kulkarni2016hierarchical}.

A primary reason behind RL's sample inefficiency is the lack of prior knowledge used in training new policies. One of the biggest takeaways from advances in computer vision (CV) and natural language processing (NLP) is that priors (both architectural and parametric) are invaluable. Hence this begs the question: why should an agent learn a task from scratch? Why not use strong priors? Recent works have shown the promise of semantic priors such as auxiliary losses~\cite{jaderberg2016reinforcement}, and architectural priors such as transfer learning networks~\cite{rusu2016progressive}. However unlike passive domains such as object detection, problems in motor control and robotics offer another strong prior: morphology. Instead of learning policies from scratch on a given agent, we can use policies previously learned on morphologically different agents as a prior. For example, instead of training a quadruped walking policy to solve a maze from scratch, requiring millions of expensive samples, we could transfer-learn from a much simpler robot, say a Roomba wheeled robot. This morphological transfer affords two benefits: first, learning a policy first on a simple morphology and then transferring to a harder one induces a natural curriculum for learning~\cite{bengio2009curriculum} which provides richer rewards and learning signal; second, this allows us to use fewer samples from complex robots that are often expensive and time-consuming.

\begin{figure}[t!]
\centering
\includegraphics[width=0.8\linewidth]{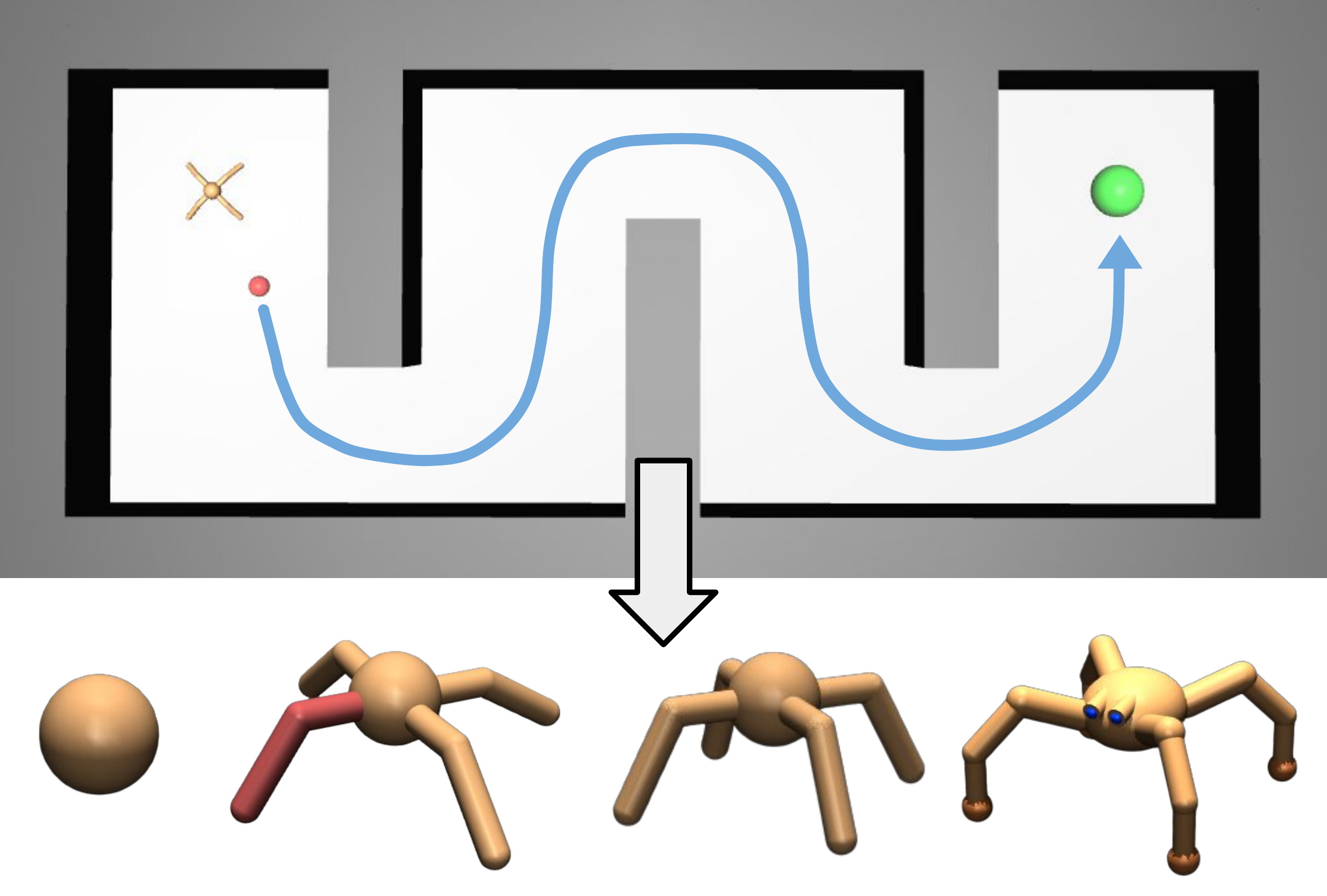}
\vspace{-0.1in}
\caption{In this work, we focus on the problem of transferring long-horizon policies across morphologies. Hence, given an already performant simple agent, we imitate its behaviors on a more complex agent for transfer through hierarchical decoupling.}
\label{fig:intro}
\vspace{-0.15in}
\end{figure}

But how do we effectively transfer policies across morphologies? Direct policy transfer is infeasible, since different morphologies have different state and action spaces (see \Figref{fig:intro} for examples of morphologies). Another option would be to use morphological latent embeddings ~\cite{hcp,pathak2019learning}, but learning robust embeddings requires training across hundreds of morphologies. Instead, inspired by recent advances in hierarchical learning~\cite{kulkarni2016hierarchical,nachum2018near}, we propose a transfer learning framework using hierarchically decoupled policies. In this framework, the low-level policy is trained specific to a given morphology, while the high-level policy can be re-used across morphologies. For compatible transfer, only the high-level policy, operating on a global agent state is transferred, while the low-level policy is independently learned. 

However, as recent work by \citet{hiro} has shown, high-level policies are intricately tied with low-level policies. Intuitively, if a low-level policy isn't able to reach a specific sub-goal requested by the high-level policy, the high-level policy will not select that sub-goal. This brings a significant challenge to morphological transfer for agents. Since the low-levels of two agents might be significantly different due to differences in morphology that afford varying low-level capabilities, a zero-shot transfer of the high-level policy might not always be successful. To counter this, we propose a top-down alignment of the low-level policies. This is done by introducing a information theoretic alignment loss that minimizes the mutual information between morphologies and low-level behaviors. This objective is practically optimized using discriminative learning, which allows the low-level policy of a more complex agent to imitate the behavior of the simpler agent's, making high-level transfer more successful.

Using this low-level alignment significantly improves the transfer of high-level policies. However, even with better alignment, zero-shot transfer of high-level policies will not be able to fully utilize the additional benefits of a complex morphology. One way to improve on this is to finetune~\cite{girshick2014rich} the high-level policies after the zero-shot transfer. But, in RL straightforward finetuning suffers from catastrophic forgetting~\cite{rusu2016progressive} of the simpler agent's high-level under the changing dynamics of the new agent. To prevent this, we take inspiration from prior work in transfer learning and introduce a KL-regularizer that allows the complex agent to improve performance while staying close to the simpler agent's high-level. Intuitively, this balances the imitation of the simpler agent's high-level with its own ability to the solve the task. Doing this significantly improves performance on a suite of navigation and manipulation transfer tasks.

In summary, we present the following contributions in this work: (a) we show how hierarchical policies can help morphological transfer. Although recent works in Hierarchical Reinforcement Learning ~\cite{peng2017deeploco,peng2019mcp} allude to the potential of morphological transfer, to our knowledge we are the first work that concretely focuses on this problem. (b) We propose two key technical insights for hierarchical imitation, a top-down low-level alignment and a KL-regularized high-level objective to accelerate transfer. (c) Finally, we empirically demonstrate significant improvements in performance of morphological transfer on long-horizon navigation and manipulation tasks.

%% file: method.tex
\section{Background}

\begin{figure*}[t!]
\centering
\includegraphics[width=\textwidth]{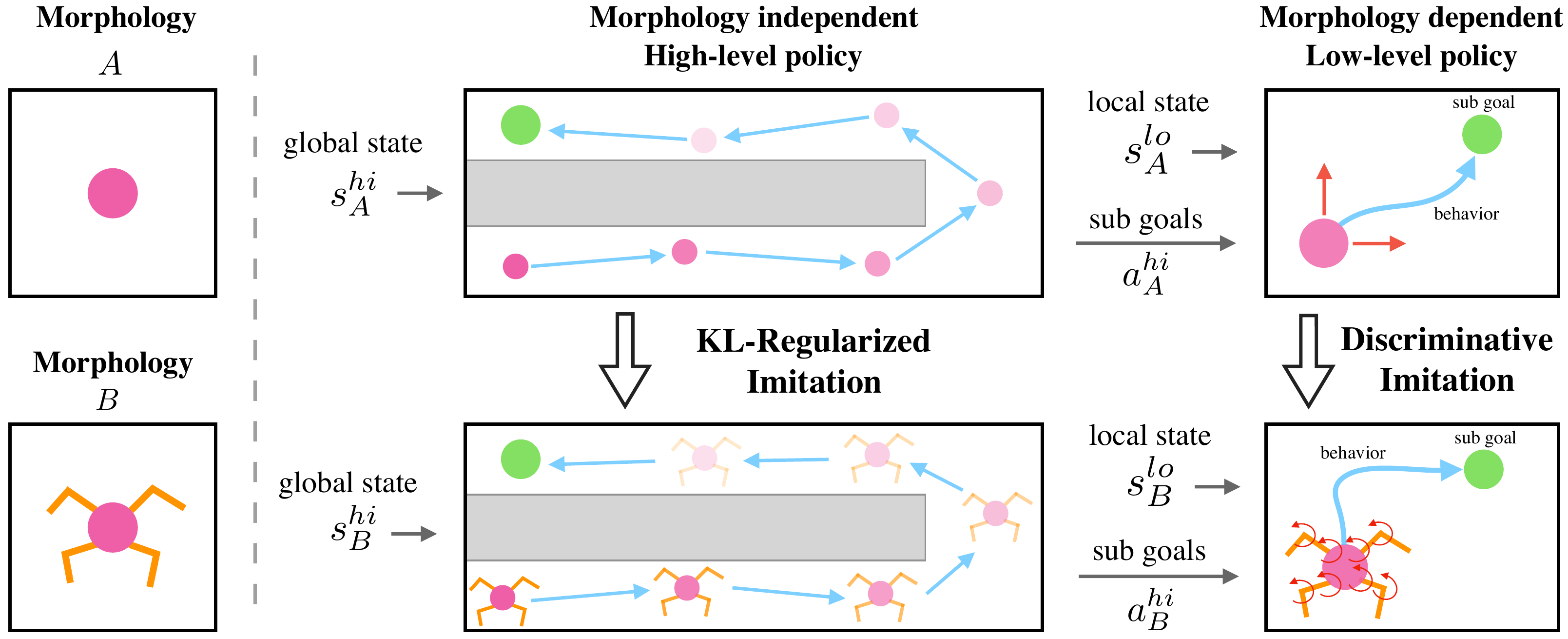}
\caption{We transfer Hierarchical policies across morphologies through decoupled imitation. For the low-level policy we use density based discriminative imitation detailed in \Secref{sec:disc}, which improves zero-shot high-level transfer. For the high-level policy, we use KL-regularized imitation detailed in \Secref{sec:kl_reg}, which improves finetuning the high-level.}
\label{fig:training}
\end{figure*}

Before we describe our framework, we first discuss relevant background on RL. For an in-depth survey, we refer the reader to~\cite{sutton1998introduction,kaelbling1996reinforcement}.

\subsection{Reinforcement Learning}
In our continuous-control RL setting, an agent receives a state observation $s_t \in \mathcal{S}$ from the environment and applies an action $a_t \in \mathcal{A}$ according to policy $\pi$. With stochastic policies, as in our case, we have $a_t \sim \pi(s_t)$. The environment returns a reward for every action $r_t$. The goal of the agent is to maximize expected cumulative discounted reward $\expect_{s_{0:T},a_{0:T-1},r_{0:T-1}}\left[\sum_{t=0}^{T-1} \gamma^t r_t\right]$ for discount factor $\gamma$ and horizon length $T$. On-policy RL ~\cite{schulman2015trust,kakade2002natural,williams1992simple} optimizes $\pi$ by iterating between data collection and policy updates. It hence requires new on-policy data every iteration, which is expensive to obtain. On the other hand, off-policy reinforcement learning retains past experiences in a replay buffer and is able to re-use past samples. Thus, in practice, off-policy algorithms have been found to achieve better sample efficiency \cite{lillicrap2015continuous}. For our experiments we use the off-policy SAC~\cite{haarnoja2018soft} algorithm as our base RL optimizer due to its sample efficiency. However, our framework is compatible with any standard RL algorithm.

\subsection{Two-layered Hierarchical RL (HRL)}
\label{sec:hrl_setup}
The central idea of HRL is to abstract the policy $\pi$ into multiple policies that operate at temporally different levels. The most common abstraction is a two-level hierarchy~\cite{hiro}, a low-level policy $\pi^{lo}$ and a high-level policy $\pi^{hi}$. In our formulation, the high-level takes as input a part of the state observation $s^{hi}_t$ and outputs morphology-independent high-level actions $a^{hi}_t$ that serve as subgoals for the low-level policy. These subgoals are morphology-independent and belong to a goal space $\sG$. The low-level takes as input a part of the state observation $s^{lo}_t$ and the commanded subgoal $a^{hi}_t$, and outputs low-level control actions $a^{lo}_t$ to try and reach the subgoal. Note that we split the state observation $s_t$ into two components, $s^{hi}_t$ a global morphology-independent state and $s^{lo}_t$ a proprioceptive morphology dependent state. This particular choice of state, inspired from \citet{marino2018hierarchical}, allows us to transfer $\pi^{hi}$ across morphologies that have different state representations.



\section{Method}

In this work, we focus on the problem of transferring policies from one morphology to another for challenging long-horizon problems. Concretely, we define morphology transfer from morphology $A$ to morphology $B$ on a specific task as transferring a policy trained with $A$ to $B$, i.e., given access to a trained $\pi_A$, what is the fastest way to train $\pi_B$ (See \Figref{fig:training}). Practically, $A$ is a simple agent for which training $\pi_A$ might be a lot easier (and safer) than training on a more complex agent $\pi_B$. However, since $A$ and $B$ have different state and action spaces afforded by their morphologies, we build on top of the hierarchical setup described in \Secref{sec:hrl_setup}. The hierarchical demarcation between low-level policies that act on proprioceptive states and high-level policies that act on global states provides a natural way to transfer high-level knowledge. Moreover, hierarchical learning is empirically known to provide massive improvements in sample-efficiency~\cite{hiro}. In the following subsections we detail our proposed technique.

\subsection{Zero-shot High-level Transfer}
\label{sec:zero_shot}
In the hierarchical setting, morphology transfer reduces to transferring  $\pi_A \equiv [\pi^{lo}_A, \pi^{hi}_A,]$ trained on $A$ to $\pi_B \equiv [\pi^{lo}_B, \pi^{hi}_B,]$. One straightforward way to perform transfer is to set $\pi^{hi}_A \rightarrow \pi^{hi}_B$ as the input $s^{hi}$ and output $a^{hi}$ of the high-level policies are morphology-independent. The low-level policy $\pi^{lo}_B$ can either be learned with the fixed high-level $\pi^{hi}_B$ or trained independently. In our experiments, we pre-train the low-level policy $\pi^{lo}_B$ on uniformly sampled goals from $\sG$, allowing us to learn an effective $\pi^{lo}_B$ without access to $\pi_A$ and generalize over tasks without re-training $\pi^{lo}_B$.



\subsection{Low-level Imitation Through Behavior Alignment}
\label{sec:disc}

Although directly transferring the high-level policy $\pi^{hi}_A \rightarrow \pi^{hi}_B$ with independently trained low-levels $\pi^{lo}_B$ allows for zero-shot morphology transfer, it suffers from low-level domain shift especially for tasks requiring precise control. This is because different morphologies afford different low-level behavior. Intuitively, if $\pi^{lo}_B$ doesn't generate similar behavior to $\pi^{lo}_A$, transferring $\pi^{hi}_A \rightarrow \pi^{hi}_B$ may not work since $\pi^{lo}_B$ does not generate the behavior expected by  $\pi^{hi}_A$. To solve this problem, we align the low-level $\pi^{lo}_B$ to $\pi^{lo}_A$ on the set of goals $\sG$. Note that direct cloning is not possible since the low-level state and action spaces are not the same. Additionally, simply ensuring both agents can reach the same portions of $\sG$ used by $\pi^{hi}_A$ is insufficient for strong alignment since goal-reaching behavior can differ while achieving the same goal (illustrated as low-level behavior in \Figref{fig:training}). Even in the unlikely scenario where $\pi_A^{lo}$ and $\pi_B^{lo}$ can reach the same subset of $\sG$ within a single high-level step, different inter-step trajectories can still affect non-static environments.

To incentivize $\pi^{lo}_B$ (parameterized by $\theta^{lo}_B$) to mimic the behavior of $\pi^{lo}_A$, we propose minimizing the following mutual information objective:
\begin{equation}
\label{eq:info}
    \min_{\theta^{lo}_B} I(\textrm{morphology}; \textrm{behavior})
\end{equation}
Minimizing the mutual information between the morphology type ($M=\{A, B\}$) and the generated low-level behavior will result in a low-level policy $\pi^{lo}_B$ whose behavior cannot be distinguished from $\pi^{lo}_A$. To ensure that the behavior is compatible with both morphologies $A$ and $B$, we set behavior at time $t$ as $\tau_t = s^{hi}_{t-k:t}$, where $k$ is the horizon of behavior. Let $\Tau_t$ denote the distribution over behaviors $\tau_t$.

As $I(M;\Tau_t) = H(M) - H(M|\Tau_t)$ and $H(M)$ cannot be controlled through $\pi^{lo}_B$, our objective from \eqref{eq:info} reduces to maximizing $H(M|\Tau_t)$. Here $H(\cdot)$ denotes entropy. Since the probability of agent selection $P(M=m | \Tau_t = \tau_t)$ cannot be readily estimated, we compute the variational lower bound, which reduces our objective to:
\begin{equation}
\label{eq:vlb}
    \max_{\theta^{lo}_B} \E_{m \sim M, \tau_t \sim \Tau_t}[-\log q_\phi(m | \tau_t)]
\end{equation}
Here $q$ parameterized by $\phi$ is effectively a binary classifier (or discriminator) that outputs the probability of the generated behavior coming from morphology $m$ given input behavior $\tau_t$. Complete derivations of \eqref{eq:vlb} can be found in Appendix A. Maximizing this objective through $\theta^{lo}_B$ implies generating behaviors that maximally confuse the discriminator. To do this we augment the low-level policy rewards as follows:
\begin{equation}
\label{eq:LL_update}
    r^{lo}_B \leftarrow R(\tau^B_t| g) - \lambda_0^{\lambda_1} \log q_\phi(M=B| \tau^B_t)
\end{equation}
Here $\lambda_0$ and $\lambda_1$ represent temperature parameters that anneals the rewards from the discriminator over time, $\tau^B_t$ represents the trajectory generated by $\pi^{lo}_B$ while trying to solve the subgoal $g$, and $R$ represents the sub-goal reward function. The training process is summarized in \Algref{alg:adv_rl}. This procedure of fitting a discriminator and then re-optimizing the policy has a similar flavor to recent work in approximate inverse reinforcement learning~\cite{ho2016generative}, where the discriminator represents the reward density function.

\begin{algorithm}[tb]
   \caption{Low-level Alignment for high-level transfer}
   \label{alg:adv_rl}
\begin{algorithmic}
   \STATE {\bfseries Input:} New agent $B$, agent $A$'s behavior $\Tau_A$, and a goal distribution $\sG$. 
   \STATE {\bfseries Initialize:} Learnable parameters $\theta$ for $\pi^{lo}_{B,\theta}$ and $\phi$ for $q_\phi$
  \FOR {$i$=1,2,..$N_\text{iter}$}
  \STATE sample goal $g \sim \sG$
  \FOR {$j$=1,2,..$T$}
  \STATE sample action $a^{lo}_t \sim \pi^{lo}_{B,\theta}$
  \STATE collect experience $(s^{lo}_t, a^{lo}_t, s^{lo}_{t+1}, r^{lo}_t)$ for $\Tau_B$
  \ENDFOR
  \FOR {$j$=1,2,...$M_\text{policy}$}
  \STATE update $r^{lo}_t$ according to eq \ref{eq:LL_update}
  \STATE $\theta \leftarrow \text{policyOptimizer}(\{(s^{lo}_t, a^{lo}_t, s^{lo}_{t+1}, r^{lo}_t)\}, \theta)$
  \ENDFOR
  \FOR {$j$=1,2,...$M_\text{discrim}$}
  \STATE $\phi \leftarrow \text{discrimOptimizer}(\Tau_A, \Tau_B, \phi)$
  \ENDFOR
  \ENDFOR
  \STATE {\bf Return: }{$\theta$}
\end{algorithmic}
\end{algorithm}

\subsection{High-level Imitation Through KL-regularized Training}
\label{sec:kl_reg}

In the previous section we discussed aligning the low-levels through density based imitation, which allows for better zero-shot transfer of the high-levels $\pi^{hi}_A \rightarrow \pi^{hi}_B$. However, even with low-level alignment, if the morphologies afford different abilities, direct transfer of the high-level may not reach optimal behavior. One way to remedy this is to finetune the high-level $\pi^{hi}_B$ after transferring from $\pi^{hi}_A$, which already encodes knowledge required to solve the task. However, direct finetuning with RL often leads to catastrophic forgetting of the former policy $\pi^{hi}_A$ as previously noted in ~\citet{rajeswaran2017learning,rusu2016progressive}. To alleviate this, we propose a KL-regularized finetuning that balances staying close to $\pi^{hi}_A$ and exploiting task driven reward signals. 
\begin{equation}
\label{eq:HL_update}
    \text{grad}_B^{hi} \leftarrow \textrm{grad}_{\text{RL}} + \alpha \nabla_{\theta^{hi}_B} \text{KL}(\pi^{hi}_B(a^{hi}_B | s^{hi}_B)|| \pi^{hi}_A(a^{hi}_A | s^{hi}_B))
\end{equation}
Here, under the high-level trajectory $\tau^{hi} \equiv (s^{hi}, a^{hi})$ generated by $\pi^{hi}_B$, the gradients for the parameters $\theta^{hi}_B$ of $\pi^{hi}_B$ are represented by $\text{grad}^{hi}_B$. $\text{grad}_{\textrm{RL}}$ represents the gradients through the base RL optimizer. While the first term of equation \ref{eq:HL_update} gives the RL gradient, the second term represents the behavior cloning gradient of imitating $\pi^{hi}_A$, with $a^{hi}_A \sim \pi^{hi}_A$. This additional imitation objective is a practical form of regularization that penalizes deviations from sub-goal sequences set by the existing high-level $\pi^{hi}_A$. Additionally, the imitation loss forces the policy to remain near the portion of the state space in which $\pi^{hi}_A$ was maximally performant, preventing the aforementioned phenomena of catastrophic forgetting.

%% file: results.tex
\newcommand{\vsarr}[1][3pt]{\mathrel{%
   \hbox{\rule[\dimexpr\fontdimen22\textfont2-.2pt\relax]{#1}{.4pt}}%
   \mkern-4mu\hbox{\usefont{U}{lasy}{m}{n}\symbol{41}}}}

\begin{figure*}[t!]\centering
\includegraphics[width=0.85\linewidth]{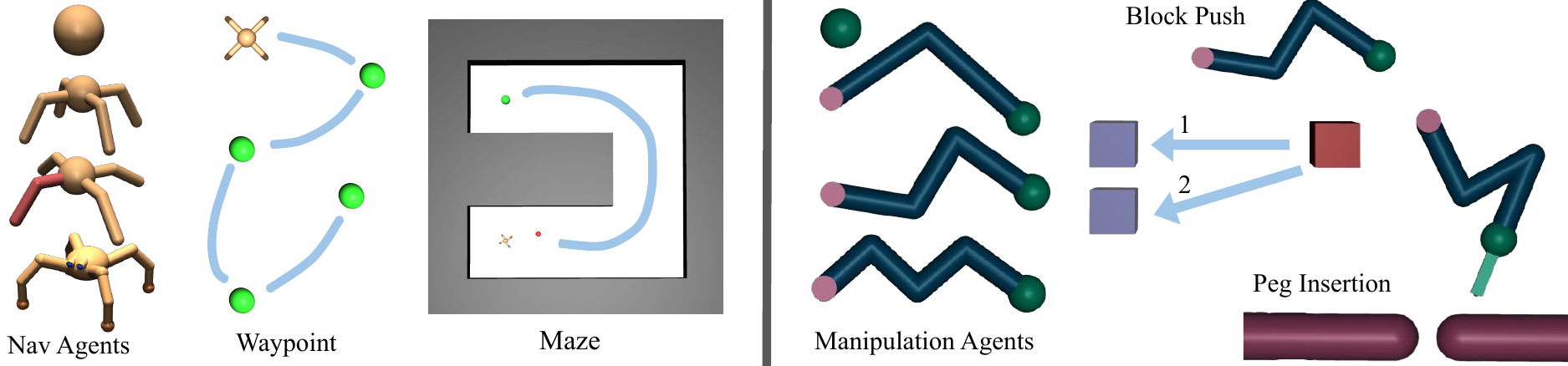}
\caption{Visualization of all agents and tasks used in our environment suite.}
\label{fig:tasks}
\end{figure*}

\section{Experiments}
In this section we discuss empirical results on using hierarchically decoupled imitation for transferring policies across morphologies. Note that since there are no standard benchmarks for evaluating morphological transfer, we create an environment suite that will be publicly released. Using this experimental setup, we seek to answer the following questions. First, does hierarchical decoupling provide an effective natural framework for morphology transfer? Second, does discriminative imitation improve zero-shot performance of morphology transfer? Finally, does KL-regularized fine-tuning improve the performance of transferred policies over baseline methods?

\subsection{Experimental Setup}
To study morphological transfer for long-horizon tasks, we present a suite of environments with eight agents and four tasks for manipulation and navigation as illustrated in \Figref{fig:tasks}. For navigation, we use four agents: \textit{PointMass}, \textit{Ant}~\cite{duan2016benchmarking}, \textit{3-Legged Ant}, \textit{Quadruped}~\cite{tassa2018deepmind}, and two tasks: \textit{Waypoint} navigation and \textit{Maze} navigation. Note that unlike in \citet{hiro}, our maze task offers only a sparse reward. During high-level training we sample goals uniformly across the six segments of the maze, but in evaluation we consider both \textit{Maze Sample}, where goals are sampled as in training and \textit{Maze End}, where the goal is always set to the end of the maze. All morphologies have vastly different action spaces. The simplest agent, the \textit{PointMass}, has an action space of two dimensions while the most complex agent, the \textit{Quadruped},  has an action space of twelve dimensions. Across navigation tasks, the goal space is set to the $(x,y)$ of the agent's torso. For the manipulation environments, we use four agents with varying degrees of freedom: \textit{PointMass}, \textit{2Link-Arm}, \textit{3Link-Arm}, and \textit{4Link-Arm} adapted from the OpenAI Gym Reacher \cite{brockman2016openai} and two tasks: \textit{BlockPush} (with variants based on goal location) and \textit{PegInsert}. The goal space $\sG$ for the manipulation tasks is the $(x,y)$ position of the end-effector. All of these environments are simulated in MuJoCo \cite{todorov2012mujoco} using the OpenAI Gym interface. Images of the environments can be found in Figure \ref{fig:tasks}. More details of the environments and the agents are provided in Appendix C.

\subsection{Training Details}
We use SAC as our base RL optimizer, and re-purpose the open-source Stable Baselines~\cite{stable-baselines} code-base for our methods. We first train low-level policies by sampling uniformly from the goal space of each task $\sG$. In order to make our hierarchical framework applicable to any task or environment, we use reward functions for the low-level policies that are highly generalizable. For the navigation tasks, low-level reward is given by the weighted cosine between the vector of agent movement and vector to the goal $g$. In all manipulation tasks, we use $L2$ distance between the end-effector and goal $g$ with a sparse reward for being within $\epsilon$ of $g$. Across all tasks of the same type, we only train low-level policies for each agent once; i.e. the same low-level policies used for \textit{Waypoint} navigation are used for the \textit{Maze} task. We ran all experiments across five random seeds, except for the high-level and high-level finetunings of the selected navigation tasks, where we ran ten random seeds.

When training policies for low-level imitation we collect goal-conditioned transition data in the form $(g - s^{hi}_t, g - s^{hi}_{t+1})$ from both agents in an off-policy manner. Data from agent $A$ is collected by running $\pi_A^{lo}$ on uniformly sampled goals from $\sG$, however one could imagine re-using previous data generated while learning $\pi_A^{lo}$. Data from agent $B$ is reused from the RL algorithm. For every step of agent $B$ during optimization, we add its transition along with a transition from agent $A$ to a circular buffer that maintains class balance. While training, we only update the discriminator periodically by randomly sampling data from the buffer. In addition to annealing the weight of discriminator rewards as per equation \ref{eq:LL_update}, we also anneal the learning rate of the discriminator to zero, preventing over-fitting and allowing agent $B$ to train against an increasingly stationary target.

For KL-regularized finetuning, we directly incorporate a term for the KL-divergence between $\pi_A^{hi}$ and $\pi_B^{hi}$ into the policy optimizer loss. This is easily calculated as all of our policies are parameterized by diagonal gaussians. Just as with the discriminator, the KL-loss coefficient is annealed to zero during training as it is no longer needed once the transferred policy is stable. Additional training details including hyper-parameters set are included in Appendix D.

\subsection{How well does zero-shot hierarchical transfer work?}


\begin{table*}[]
\centering
\resizebox{\textwidth}{!}{
\begin{tabular}{l|llll||l|ll||l|ll}
Waypoint & PM HL               & Ant HL       & Ant3 HL      & Quad HL      & Maze E   & PM HL         & Ant HL        & Maze S & PM HL         & Ant HL        \\ \hline
PM LL    & $1\text{e}3 \pm 43$ & $603 \pm 34$ & $716 \pm 58$ & $577 \pm 70$ & PM LL   & $.80 \pm .18$ & $.19 \pm .17$ & PM LL        & $.96 \pm .04$ & $.56 \pm .04$ \\
Ant LL   & $483 \pm 39$        & $476 \pm 19$ & $473 \pm 51$ & $407 \pm 36$ & Ant LL  & $.30 \pm .08$ & $.16 \pm .14$ & Ant LL       & $.62 \pm .04$ & $.50 \pm .08$  \\
Ant3 LL  & $489 \pm 74$        & $484 \pm 72$ & $499 \pm 65$ & $432 \pm 65$ & Ant3 LL & $.58 \pm .13$ & $.12 \pm .10$ & Ant3 LL      & $.84 \pm .04$ & $.56 \pm .05$ \\
Quad LL  & $169 \pm 33$        & $182 \pm 22$ & $220 \pm 27$ & $257 \pm 19$ & Quad LL & $.40 \pm .21$ & $.13 \pm .15$ & Quad LL      & $.78 \pm .08$ & $.58 \pm .04$            
\end{tabular}
}
\vspace{0.08in}\\
\resizebox{0.87\textwidth}{!}{
\begin{tabular}{l|llll||l|lll}
Block Push 1 & PM HL & 2Link HL & 3Link HL & 4Link HL & Insert   & PM HL & 3Link HL & 4Link HL \\ \hline
PM LL    &$.99 \pm .01$&$.21 \pm .15$&$.33 \pm .18$&$.26 \pm .17$& PM LL    &$1.0 \pm .00$&$.60 \pm .22$&$.60 \pm .22$\\
3Link LL &$.17 \pm .08$&$.20 \pm .09$&$.96 \pm .02$&$.49 \pm .14$& 3Link LL &$1.0 \pm .00$&$1.0 \pm .00$&$.80 \pm .18$\\
4Link LL &$.24 \pm .11$&$.07 \pm .04$&$.10 \pm .05$&$.89 \pm .04$& 4Link LL &$1.0 \pm .00$&$.80 \pm .18$&$.90 \pm .06$
\end{tabular}
}
\caption{Selected Zero-Shot performance results averaged across a hundred episodes per run with navigation tasks on top and manipulation tasks below. For all tasks except waypoint navigation, values reported indicate the fraction of successes. ``Maze E" refers to the \textit{Maze End} evaluation and ``Maze S" refers to the \textit{Maze Sample} evaluation.}
\label{tab:zeroshot}
\end{table*}

We perform straightforward high-level transfer described in \Secref{sec:zero_shot} across our task-morphology environment suite. In Table \ref{tab:zeroshot}, we present results from combining the high-level of an agent (column-wise) with a specific agent (row-wise). On relatively easy tasks like \textit{Waypoint}, zero-shot transfer works well. For instance, on the \textit{Ant} morphology, using a high-level learned on \textit{PointMass} achieves the same performance within confidence bounds. On the sparse \textit{Maze} task, transferring from \textit{PointMass} to \textit{Ant} does better than learning the \textit{Ant} policy from scratch. Though initially surprising, this result can be explained by the \textit{PointMass}'s superior exploration--the \textit{PointMass} is speedy allowing it to easily discover sparse reward signals, while the \textit{Ant} is slow and topples over when it runs into walls. This is a concrete example of how a simple agent can provide a valuable learning curriculum to a complex one. However, the 30\% \textit{MazeEnd} success rate after zero-shot transfer still leaves much to be desired in comparison to the \textit{PointMass}'s 80\% success rate. Interestingly, even on the simple \textit{Waypoint} task the nearly ideal \textit{PointMass} experiences a significant performance degradation when using a different high-level policy, indicating that high-level policies are indeed overfit to the morphology they are trained on. This is manifested further in the \textit{BlockPush} task where zero-shot performance deteriorates significantly. For example, when transferring the high-level form the \textit{PointMass} to the \textit{4Link Arm}, performance drops by around $74\%$. The poor high-level transfer on harder environments and morphologies motivates the need for better transfer algorithms. Additional results are presented in Appendix E.


\subsection{Does discriminative imitation improve transfer?}

\begin{table*}[]
\centering
\resizebox{0.9\textwidth}{!}{
\begin{tabular}{l|lll|ll|ll|l}
Task           & \multicolumn{3}{l|}{Waypoint}      &\multicolumn{2}{l|}{Maze End}  & \multicolumn{2}{l|}{Maze Sampled} & Insert \\ \hline
Transfer       & PM$\vsarr$Ant& PM$\vsarr$Ant3 & PM$\vsarr$Quad & PM$\vsarr$Ant & PM$\vsarr$Quad & PM$\vsarr$Ant      & PM$\vsarr$Quad & 3Link$\vsarr$4Link \\ \hline
Zero-Shot      &$483 \pm 39$&$489 \pm 74$&$169 \pm 33$&$.30 \pm .08$&$.40 \pm .22$&$.62 \pm .04$&$.78 \pm .08$&$.80 \pm .18$\\
Discrim (ours) &$546\pm33$  &$495\pm38$  &$338\pm 39$ &$.55 \pm .13$&$.40 \pm .21$&$.72\pm .03$ &$.89 \pm .06$&$.93 \pm 0.05$
\end{tabular}
}
\vspace{0.08in}\\
\resizebox{\textwidth}{!}{
\begin{tabular}{l|llll|llll}
Task           & \multicolumn{4}{l|}{Block Push 1}                      & \multicolumn{4}{l}{Block Push 2}                       \\ \hline
Transfer       & PM$\vsarr$3Link & 2Link$\vsarr$3Link & 2Link$\vsarr$4Link & 3Link$\vsarr$4Link & PM$\vsarr$3Link & 2Link$\vsarr$3Link & 2Link$\vsarr$4Link & 3Link$\vsarr$4Link  \\ \hline
Zero-Shot      &$.17 \pm .08$ &$.20 \pm .09$&$0.07\pm .04$&$.10 \pm .05$&$.24 \pm .12$ &$.61 \pm .16$&$.19 \pm .12$&$.39 \pm .16$ \\
Discrim (ours) &$.34 \pm .10$ &$.43 \pm .15$&$.17 \pm .08$&$.23 \pm .13$&$.43 \pm .12$ &$.42 \pm .11$&$.46 \pm .30$&$.44 \pm .14$
\end{tabular}
}

\caption{Selected discriminator results. We train low-level policies with a discriminator for the transferred-to agents, then assess zero-shot performance across a hundred episodes for each run. We find that the same low-level aligned policy improves transfer across both tasks, commonly leading to a doubling of performance in the \textit{Block Push} tasks.}
\label{tab:disc}
\end{table*}

To improve zero shot high-level transfer, we perform discriminative imitation on the low-level policies as described in \Secref{sec:disc}. Results are presented in table \ref{tab:disc} across a wide set of poorly performing plain high-level transfers. We measure the effectiveness of the discriminator by comparing zero-shot performance of the plain low-level and the low-level trained with a discriminator for a given high-level. Across nearly all tasks, discriminative imitation of the low-level improves zero-shot performance, especially in manipulation tasks. For example, transferring the \textit{PointMass} high-level to the \textit{3Link Arm} is twice as successful across both block push tasks using our method. The \textit{Quadruped}'s performance in the \textit{Waypoint} task with discriminative imitation was better than training the high-level from scratch for the limited number of low-level steps we used in training. A substantial benefit of discriminative imitation is that the aligned low-level policy only needs to be trained once for transfer across any number of high-level policies, making it particularly useful when training the high-level policy is expensive or an increase in performance is desired across a large set of tasks. Besides the \textit{2Link} to \textit{3Link} transfer, the same imitated low level boosts performance on all tasks.

\begin{figure}[t!]
\centering
\subfigure{\includegraphics[width=0.49\linewidth]{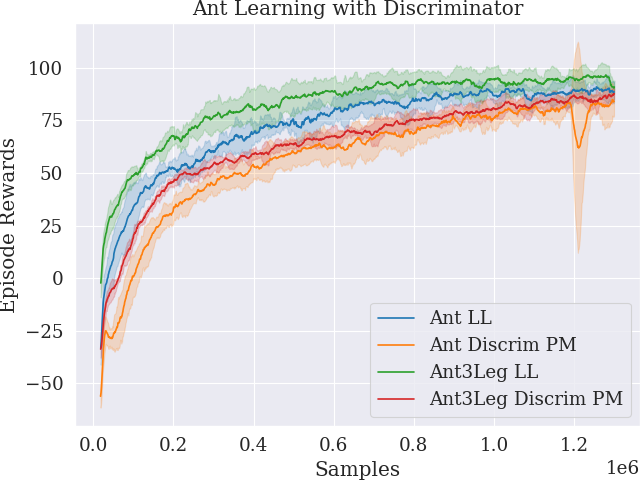}}
\subfigure{\includegraphics[width=0.49\linewidth]{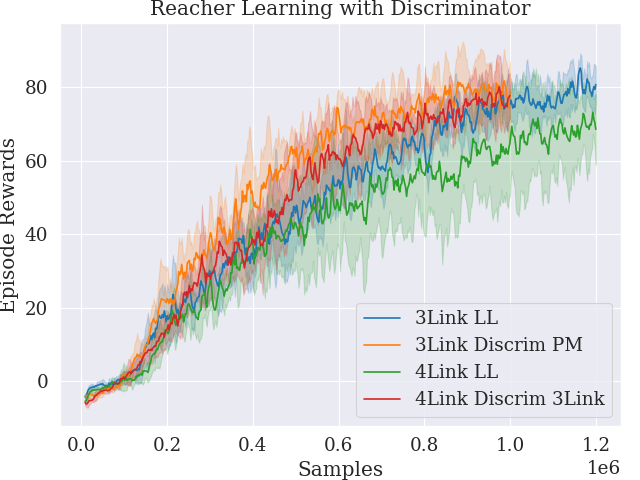}}
\caption{Learning curves for low-level policies with and without the discriminator. For the Ant Agent (left) the discriminator policy achieves slightly lower though comparable performance to the regular low-level, while the 3 and 4 Link arms (right) learn slightly faster with discriminator supervision.}
\label{fig:discrimlow}
\end{figure}

To verify that our agents are indeed learning to discriminate rather than just learning objectively better policies, we examine the low-level learning curves in \Figref{fig:discrimlow}. Though for the arm agents the discriminator provides an extra form of supervision that speeds up learning, final rewards for agents with the discriminator are near those of agents without, indicating that the performance benefits in Table \ref{tab:disc} can indeed be attributed to imitation.

\subsection{Does finetuning improve transfer?}
After transferring the high-level policy from a simpler agent to a more complex one, we finetune it by retraining to improve performance. Results for this are visualized in \Figref{fig:finetune_nav} and \Figref{fig:finetune_man} where we examine returns when using extra samples collected from agent $B$. In most cases, finetuning works better than learning the high-level from scratch (in green) and reaches substantially higher performance compared to the zero-shot high-level transfer (dotted purple line). However, in some cases regular finetuning suffers from unstable training. For instance in the transfer finetuning of \textit{2Link Arm} from \textit{PointMass} on \textit{BlockPush}, we see poor confidence bounds. Empirically, we notice massive fluctuations in training performance across seeds that cause this behavior, most likely explained by high-level policies shifting far out of distribution during morphology transfer, causing catastrophic forgetting of task-solving knowledge. This motivates the need for better transfer learning methods for finetuning.

\subsection{How much does KL-regularized training help?}
\begin{figure*}[t!]
\centering
\includegraphics[width=\linewidth]{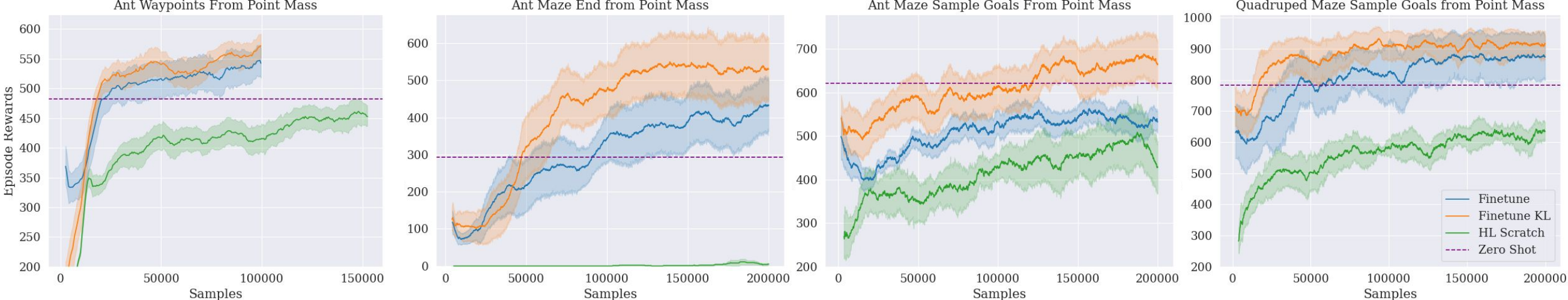}
\vspace{-0.15in}
\caption{Comparison of performance finetuning from \textit{PointMass} for \textit{Ant Waypoint}, \textit{Ant Maze End}, \textit{Ant Maze Sample}, and \textit{Quadruped Maze Sample} from left to right. For \textit{Waypoint}, the agent receives a reward of 100 per waypoint reached and for \textit{Maze}, the agent receives a reward of 1000 for reaching the set goal. ``Full" denotes that the policy was trained without hierarchy. For \textit{Maze End}, we finetune with the goal always set to the end of the maze.}
\label{fig:finetune_nav}
\end{figure*}

\begin{figure*}[t!]
\centering
\includegraphics[width=\linewidth]{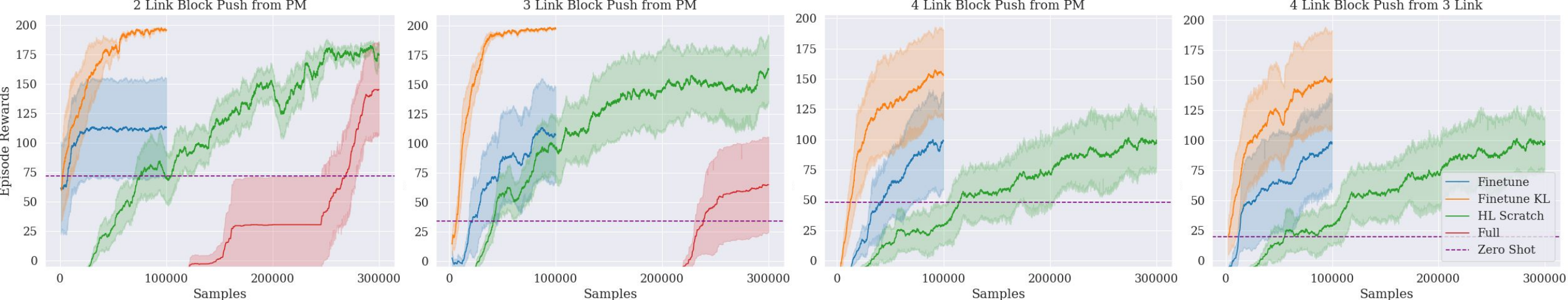}
\vspace{-0.15in}
\caption{Learning curves of finetuning for \textit{2 Link} from \textit{PointMass}, \textit{3 Link} from \textit{PointMass}, \textit{4 Link} from \textit{PointMass}, and \textit{4 Link} from \textit{3 Link} from left to right on the block push 1 task. The agent receives a reward of 200 for completing the task successfully.}
\label{fig:finetune_man}
\end{figure*}

To improve performance during finetuning, we use KL-regularized imitation as described in \Secref{sec:kl_reg}. Empirically, we find that the addition of an imitation loss during high-level transfer substantially improves performance as seen in \Figref{fig:finetune_nav} and \Figref{fig:finetune_man}. On the \textit{BlockPush} we notice at least a 2X speedup compared to already strong direct finetuning method. On the \textit{Maze} environments, we again notice improvements in performance; however the gains are modest compared to the \textit{BlockPush} environment.  We additionally compare our KL-regularized finetuning method to training the high-level policy from scratch with a pre-trained low-level and learning the task without hierarchy, denoted ``Full". Note that we make our comparisons with respect to the number of samples used in training. Learning without hierarchy was unsuccessful for the complex agents on navigation tasks, particularly the sparse-reward maze. Though it is cut off in the graphs, learning without hierarchy was eventually successful for all the manipulation tasks.

\begin{figure}[t!]
\centering
\includegraphics[width=\linewidth]{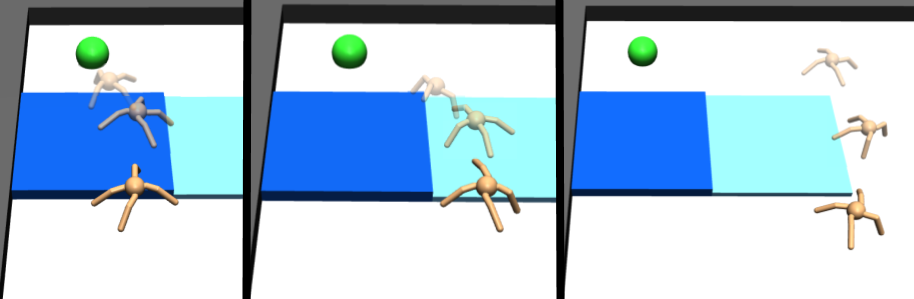}
\caption{Trajectories for the \textit{Ant} agent on a ``steps" environment for training the high-level from scratch, finetuning from \textit{Point Mass}, and zero-shot transfer from \textit{Point Mass} from left to right respectively. The dark blue (taller) and light blue (shorter) steps are reduced in height to allow the ant agent to climb over.}
\label{fig:steps}
\end{figure}

\subsection{In what cases does hierarchical transfer fail?}
\begin{table}[]
\centering
\resizebox{0.48\textwidth}{!}{
\begin{tabular}{l|ll||ll}
Zero-Shot & PM HL         & Ant HL        & Finetune    & $.53 \pm .09$ \\ \hline
PM LL     & $.60 \pm .22$ & $.00 \pm .00$ & Finetune KL & $.52 \pm .13$ \\ \cline{4-5} 
Ant LL    & $.32 \pm .16$ & $.59 \pm .05$ & HL Scratch  & $.59 \pm .05$
\end{tabular}
}
\caption{Comparison of transfer performance on the step task. We evaluate only on the goal at the end of the maze as depicted in Figure \ref{fig:steps}}
\label{tab:steps}
\vspace{-0.1in}
\end{table}

The aforementioned results demonstrate the significant promise of hierarchically decoupled transfer in scenarios where all agents can similarly cover the task state space. However, what happens if a more complex agent's morphology endows it with additional abilities? In this specific context, hierarchical decoupling may not lead to a perfectly optimal transfer as the more complex agent may be able to reach new states other agents could not. To examine such scenarios, we design a simple ``step maze" task, with low barriers that an \textit{Ant} could scale but a \textit{Point Mass} could not. We then examine the trajectories of the \textit{Ant} when its high-level is trained from scratch, finetuned form \textit{Point Mass}, and zero-shot transferred from \textit{Point Mass} in \Figref{fig:steps}. The \textit{Ant} high-level trained from scratch learns to climb over the highest barrier while the zero-shot trajectory mostly avoids the barriers just as the \textit{Point Mass} would. By finetuning, the \textit{Ant} agent moves closer to the policy learned from scratch climbing the lower barrier, but remains somewhat tied to its prior and less consistently climbs the high barrier. Empirical results are included in Table \ref{tab:steps}, where we see that for a limited number of samples from \textit{Ant}, training from scratch achieves the highest performance unlike in previous environments. By training from scratch, the \textit{Ant} reaches the final goal $59\%$ of the time while zero-shot and finetuning are only successful $32\%$ and $53\%$ of the time respectively. Furthermore, KL-regularized finetuning does not seem to help here, indicating that high-level imitation is only useful when the optimal solution for a task is similar between agents.

%% file: related_work.tex
\section{Related Work}
Our work is inspired by and builds on top of a broad range of topics, including transfer learning, morphological transfer, imitation learning, and information theoretic RL. In this section, we overview the most relevant ones. 



\subsection{Multi-task and Transfer Learning}
Learning models that can share information across tasks has been concretely studied in the context of multi-task learning ~\cite{caruana1997multitask}, where models for multiple tasks are simultaneously learned. More recently, \citet{kokkinos2017ubernet, doersch2017multi} looks at shared learning across visual tasks, \citet{pinto2017learning} looks at shared learning across robotic tasks, and \citet{pathak2019learning} looks at message passing for low-level control of articulated agents. In contrast to multi-task learning where knowledge is simultaneously learned, we focus on the disparate setting in which knowledge from one task (a simpler agent) is transferred to another (a more complex agent). \citet{murali2018cassl} learn manipulation policies via cirriculum over joints, but focus only on a single agent. More relevant to our work, \citet{hcp} focuses on transfer across different hardware by training hardware conditioned embeddings across a large number of morphologies. However, access to more than two or three morphologies is unrealistic in practice. Our method is best suited for these scenarios.

Transfer learning~\cite{pan2009survey,torrey2010transfer} focuses on transferring knowledge from one domain to another. One of the simplest forms of transfer is finetuning~\cite{girshick2014rich} using models initialized on different tasks. Several other works look at more complex forms of transfer~\cite{yang2007adapting,hoffman2014continuous,aytar2011tabula,saenko2010adapting,kulis2011you,fernando2013unsupervised,gopalan2011domain,jhuo2012robust,ammar2015unsupervised}. The most relevant to our work is \citet{tzeng2017adversarial}, where a discriminator is used to align intermediate features across domains. Similar in spirit, in our proposed low-level alignment through density imitation, we use a discriminator to align the behavior of low-level policies across morphologies.

In the context of RL, transfer learning~\cite{taylor2009transfer} research has focused on learning transferable features across tasks~\cite{parisotto2015actor,barreto2017successor,omidshafiei2017deep}.  Another line of work by \cite{rusu2016progressive,kansky2017schema,devin2017learning} has focused on network architectures that improves transfer of RL policies. Since the focus of our work is on transfer across morphologies, we note that the aforementioned works are orthogonal and complementary to ours.



\subsection{Hierarchical Reinforcement Learning}
HRL based techniques~\cite{barto2003recent,bacon2017option,li2019sub} have been able to solve complex or long-horizon tasks through temporal abstraction across hierarchies as seen in \citet{hac} and \citet{hiro}. In a similar vein, several works have focused on the discovery of primitives~\cite{diversity,DADS,shankar2020discovering}, which are useful for hierarchical RL. These ideas have already been combined, as in Stochastic Neural Networks by \citet{snn}, where skills are learned in pretraining and then used to solve diverse complex tasks. Similarly, \citet{sketches} learn modular sub-policies to solve a temporally extended task.

These prior works inform our choice to use HRL as the backbone of our framework. Moreover, hierarchical decoupling allows for a natural delineation of control~\cite{wolpert1998multiple}. However, we note that unlike standard hierarchical RL techniques, we train our low-level policies independent of the high-level policies. Although a few recent works have alluded to the potential of hierarchical policies for morphology transfer~\cite{peng2017deeploco,tirumala2019exploiting,li2019hierarchical,hu2019skill}, to our knowledge we are the first to focus on this problem.



\subsection{Imitation Learning}
The central contribution of this work is decoupled imitation learning for the low-level and high-level policies in a hierarchical setup. In the context of control, the field of learning from demonstrations (LfD)~\cite{nicolescu2003natural,argall2009survey} learns to reproduce a set of demonstrated expert behavior. A popular technique called behavior cloning~\cite{esmaili1995behavioural} focuses on fitting parametric models to expert demonstrations~\cite{kober2009learning,peters2013towards}. Several works~\cite{niekum2012learning,krishnan2018transition,murali2016tsc, meier2011movement} focus on segmenting demonstrations followed by fitting models to each of the segments. \citet{rajeswaran2017learning} focus on regularizing the behavior cloning objective for dexterous hand manipulation. Inspired by this idea, we use a KL-regularized objective in the context of regularizing the high-level policy. More  recently, \citet{goyal2019infobot} use KL-regularization in the context of goal-conditioned RL and \citet{galashov2019information} use KL-regularization in conjunction with a divided state space for learning re-usable behaviors. Additionally, \citet{sharma2019third} address the third person imitation problem, where the demonstrator is different than the acting agent, by using a hierarchical setup. 
Instead of imitation through cloning, inverse RL~\cite{ng2000algorithms,abbeel2004apprenticeship} focuses on recovering the underlying reward function from expert demonstrations. \citet{ho2016generative} extends inverse RL to higher dimensional state-action demonstrations by learning a parametric expert density model through discriminative learning between expert demonstration and learned behavior. Following this technique, several works have extended this idea to third person demonstrations~\cite{stadie2017third} and stochastic demonstrations~\cite{li2017infogail}. Solving the information theoretic formulation for low-level state alignment reduces to a similar discriminative learning approach. However, instead of differentiating between expert demonstrations and learned behavior, our discriminator differentiates between the simpler agent's low-level behavior and the more complex agent's low-level.

%% file: conclusions.tex
\section{Conclusions}
In this work, we have presented one of the first steps towards morphology transfer by using hierarchically decoupled imitation. This technique allows transferring complex long horizon behavior from morphologically simple agents to more complex ones in a fraction of sample complexity compared to standard RL techniques. Although this work focuses on simulated environments, we believe that this opens the door to research in morphological transfer on real robots. Moreover, although our technique for decoupled imitation is presented in the context of morphological transfer, we believe that the technique is flexible enough to be applied to general purpose imitation learning.

\subsection*{Acknowledgements}
We thank AWS for computing resources. We also gratefully acknowledge the support from Berkeley DeepDrive, NSF, and the ONR Pecase award. Finally, we thank Alex Li, Laura Smith, and the rest of the Robot Learning Lab community for their insightful comments and suggestions.

%% file: appendix.tex
\newpage
\section*{Appendix}
\appendix
\addcontentsline{}{section}{Appendix}

\section{Additional Derivations}
Below is the full derivation of the objective used to motivate low-level discriminative imitation, taking inspiration from other work based on information theoretic objectives \cite{diversity}. We start by minimizing the mutual information between morphology, $M$ and behavior, $\Tau_t$. $I$ denotes mutual information and $H$ denotes entropy.
\begin{align*}
    &\min_{\theta_B^{lo}} I(M; \Tau_t) = \max_{\theta_B^{lo}} -I(M;\Tau_t) \\
    &= \max_{\theta_B^{lo}} -(H(M) - H(M|\Tau_t)) \\
    &= \max_{\theta_B^{lo}} H(M|\Tau_t) - H(M) \\
    &= \max_{\theta_B^{lo}} H(M | \Tau_t) \\
    &= \max_{\theta_B^{lo}} \expect[-\log p(M |\Tau_t)] \\
    &\geq \max_{\theta_B^{lo}} \expect[-\log q_\phi(M |\Tau_t)]
\end{align*}
As per the above derivation we can encourage similar behavior across agents by maximizing the entropy of the morphology given a behavior. In the fourth step we assume the distribution over morphologies is uniform, and subsequently the second term is a constant that can be omitted from the optimization. The final step applies the variational lower bound \cite{barber2004information}. In practice, we try to align behavior when reaching the same goal. This can be accomplished by conditioning the original objective on the goal, $\min I(M;\Tau_t | g)$. Propagating this change through the derivation results in optimizing a goal conditioned discriminator $q_\phi(M|\Tau_t, g)$. In practice, we do this by using $(g - s_t, g - s_{t+1})$ as the discriminator input.

\section{Complete Algorithm}
Algorithm \ref{alg:complete} provides the complete algorithm for using both of our purposed methods of imitation, discriminative low-level and KL-regularized high-level, to transfer a policy from agent $A$ to agent $B$. Though this algorithm provides an overall training flow, note that experiments we ablate the two imitation components separately to better understand the performance contributions of each. More experimentation would be required to understand how both components interact in sequence. In general, we find that high-level KL-regularized finetuning is better for gaining performance on a specific task, whereas low-level discriminative imitation is better for boosting performance across a suite of tasks.

\begin{algorithm}[tb]
   \caption{Complete Transfer}
   \label{alg:complete}
\begin{algorithmic}
   \STATE {\bfseries Input:} Agents $A$ and $B$ and a goal distribution $\sG$. 
   \STATE {\bfseries Initialize:} Learnable parameters $\theta_A^{lo}, \theta_A^{hi}, \theta_B^{lo}, \theta_B^{hi}$ and $\phi$ for $q_\phi$.
   \STATE $\theta_A^{lo} \leftarrow \text{policyOptimizerLow}(\Tau_A, \theta_A^{lo}, \sG)$
   \STATE $\theta_A^{hi} \leftarrow \text{policyOptimizerHigh}(\Tau_A, \pi_A^{lo})$

  \FOR {$i$=1,2,..$N_\text{low}$}
  \STATE sample goal $g \sim \sG$
  \FOR {$j$=1,2,..$T$}
  \STATE collect experience $(s^{lo}_t, a^{lo}_t, s^{lo}_{t+1}, r^{lo}_t)$ for $\Tau_B$ from $\theta_B^{lo}$
  \ENDFOR
  \FOR {$j$=1,2,...$M_\text{policy}$}
  \STATE update $r^{lo}_t$ according to eq \ref{eq:LL_update}
  \STATE $\theta_B^{lo} \leftarrow \text{policyOptimizerLow}(\{(s^{lo}_t, a^{lo}_t, s^{lo}_{t+1}, r^{lo}_t)\}, \theta_B^{lo})$
  \ENDFOR
  \FOR {$j$=1,2,...$M_\text{discrim}$}
  \STATE $\phi \leftarrow \text{discrimOptimizer}(\Tau_A, \Tau_B, \phi)$
  \ENDFOR
  \ENDFOR
  \STATE $\theta_B^{hi} \leftarrow \theta_A^{hi}$
  \FOR {$i$=1,2,... $N_\text{high}$}
  \STATE collect experience $(s^{hi}_t, a^{hi}_t, s^{hi}_{t+k}, r^{hi}_t)$ for $\Tau_B$ from $\theta_B^{hi}$
  \STATE $\text{grad}_{\text{RL}} \leftarrow \text{policyOptimizerHigh}(\{(s^{hi}_t, a^{hi}_t, s^{hi}_{t+k}, r^{hi}_t)\}, \theta_B^{hi})$
  \STATE $\text{grad}_B^{hi} \leftarrow \textrm{grad}_{\text{RL}} + \alpha \nabla_{\theta^{hi}_B} \text{KL}(\pi^{hi}_B(a^{hi}_B | s^{hi}_B)|| \pi^{hi}_A(a^{hi}_A | s^{hi}_B))$
  \STATE $\theta_B^{hi} \leftarrow \text{update}(\theta_B^{hi}, \text{grad}_B^{hi})$
  \ENDFOR
\end{algorithmic}
\end{algorithm}

\section{Environment Details}
Below are more complete specifications of the environments used in experiments.
\subsection{Navigation}
For all navigation agents, the low-level reward is given by the weighted distance traveled towards the goal, with an action penalty term. 
$$ r_{lo} = \frac{(s_t^{hi} - s_{t-1}^{hi})\cdot(g_t - s_{t-1}^{hi})}{||g_t - s_{t-1}^{hi}||_2} - \lambda ||a_t||^2_2$$
High-level actions are taken once every 32 steps, except on the quadruped agent where it is performed every 64. The high-level goal space is defined to be the desired change in $x, y$ position of the agent's center, limited by a distance of four meters in either direction. \\
\textbf{Agents:}
All agents observe joint positions (\textit{qpos}), velocities (\textit{qvel}), and the vector to the next sub-goal. All agents besides the point mass additionally observe contact forces. All agents use torque control.
\begin{itemize}
    \item \textit{Point Mass}: A point mass agent whose actions are forces in the cardinal directions.
    \item \textit{Gym Ant}: This is the Open AI Gym \textit{Ant} agent with its gear reduced from 150 to 125. Note that this is less modification than the \textit{Ant} agent in HiRO \cite{hiro}.
    \item \textit{3 Leg Ant}: This agent is identical to the regular ant, expect one of its legs is frozen in place.
    \item \textit{2 Leg Ant}: Again identical to the Ant, expect two diagonally opposed legs are frozen in place.
    \item \textit{DM Control Quadruped}: The quadruped agent is similar to the Gym Ant, expect it has an extra ankle joint on each of it's legs, making controlling it different. We do not use the same control scheme as in DM Control, and instead give it the same observations as the \textit{Ant} agent.
\end{itemize}
\textbf{Tasks:}
\begin{itemize}
    \item \textit{Way Point Navigation}: The agent is tasked with navigating through a plane and reaching specific waypoints. As soon as the agent reaches one waypoint, another waypoint is randomly placed. The agent receives a reward for its $L2$ distance from the waypoint, and a sparse reward of 100 upon reaching the waypoint. The observation space is given by the agent's current position and the position of the waypoint. The high-level policy is trained with a horizon of 50 high-level steps.
    \item \textit{Maze}: The agent must navigate through a `U' shaped maze and reach the end. The agent only receives a sparse reward of 1000 upon reaching its final goal. During training, final goal locations are randomly sampled uniformly from the six "blocks" of the maze path, while in evaluation the final goal is always the end of the maze. The observation space is given by just the agent's current $(x,y)$ position and the position of the final goal.
    \item \textit{Steps}: The agent has to navigate through a similarly shaped structure to that of the \textit{Maze}, although only half the size. The height of the taller step is 0.3125 meters, while the height of the shorter step is 0.15625 meters. When the Ant agent is used for the step environment, it is given 16 rangefinder sensors and it's low level is pretrained on an environment with randomly placed steps.
\end{itemize}
\subsection{Manipulation}
For all manipulation tasks, low-level rewards are given by $L2$ distance to the selected sub-goal and an additional sparse reward. 
$$r_{lo} = - ||g_t - s_t^{hi}||_2 - \lambda ||a_t||^2_2 + \gamma 1\{g_t - f(s_t) < \epsilon\}$$
High-level planning is performed every 35 steps. Again, all agents use torque control.  \\
\textbf{Agents:}
\begin{itemize}
    \item \textit{Point Mass}: Identical to the previous point mass, just scaled to fit the environment.
    \item \textit{2-Link Arm}: This is the standard reacher from the Open AI Gym set of environments, with end effector collisions enabled.
    \item \textit{3-Link Arm}: A modified version of the standard 2-Link reacher with one extra degree of freedom. Each link is approximately one third the length of the arm.
    \item \textit{4-Link Arm}: A modified version of the standard 2-Link arm, created by splitting each link evenly into two more links. 
\end{itemize}
We found that the ant agents with fewer legs tended to be more stable and fell over less. \\

\textbf{Tasks:}
\begin{itemize}
    \item \textit{Block Push}: The arm agent has to push a block across the environment to a target end position. We test on variations of difficulty based on block position. Here, high-level observations include the position of the end effector and the position and velocity of the block. high-level rewards correspond to negative $L2$ distance of the block to its goal position and a sparse reward of 200 for solving the task. The high-level goal space is defined to be the desired change in the $x, y$ position of the agent's end effector, limited by a distance of 0.07 meters in either direction. We have two different variants of the block push task, \textit{Block Push 1}, where the block must be pushed just horizontally, and \textit{Block Push 2}, where the block must be pushed a shorter distance, but horizontally and vertically.
    \item \textit{Peg Insertion:} The agent now has a peg attached to it's end effector that it must insert into a hole. high-level observations include the position of the tip of the peg and the position of the end effector. high-level rewards correspond to negative $L2$ distance from the final desired insertion point and a sparse reward of 50 for solving the task. For peg insertion, the high-level goal space is given from the end of peg.
\end{itemize}

\section{Training Details}
When training low-level policies, we only reset environment occasionally after selecting a new low-level goal to allow the agent to learn how to perform well in long-horizon settings. Low level policies are trained over longer horizons than the exact number of steps in between high level actions. For high-level training on top of pre-trained low-levels, we collect samples only when the high-level policy sets a new sub-goal. We include hyper-parameters for all low level training in Table \ref{tab:low_hparams} and hyperparameters for all high level training in Table \ref{tab:high_hparams}.

When training the discriminator for low level imitation, we anneal the learning rate linearly from its initial value to zero over the first ``stop" fraction of training timesteps. This allows the agent to learn against an increasingly fixed target. Additionally, we anneal the discriminator weight in the reward function from it's initial value to 0.1 linearly over the first 90\% of training timesteps. Full parameters for the discriminators can be found in Table \ref{tab:discirm_hparams}. Additionally, we tested online and offline data collection. In offline data collection, transitions are randomly sampled from agent $A$'s low level policy. In online data collection, we align the goals of the two agents, such that we collect transitions of agent $A$ reaching goal $g$ when agent $B$ is attempting to reach the same $g$. The results presented in the main paper body are exclusively from offline data collection.

For KL-regularized fine-tuning, we use the same parameters across almost all experiments. We add the KL-divergence between Agent $B$'s policy and Agent $A$'s policy at every timestep. For the Waypoint task and all manipulation tasks, we use a KL weight coefficient of 1 in the loss, a learning rate of 0.01, and linearly anneal the weight of the KL loss to zero during the first 50\% of training. For the Maze Task, we lowered the learning rate to 0.001 and the KL loss coefficient to 0.01. We performed a search over learning rates for regular fine-tuning, and found the original learning rate of the policy tended to perform best and as such used it for comparison.

\begin{table*}[h]
\centering
\begin{tabular}{l|llllllll}
Agent & Timesteps & Learning Rate & Batch Size & Layers & Horizon & Reset Prob & Buffer Size & DM \\ \hline
PM (Nav) & 200000 & 0.0003 & 64 & 64 64 & 35 & 0.1 & 200000 & 4 \\
Ant(s) & 2500000 & 0.0008 & 100 & 400 300 & 100 & 0.1 & 1000000 & 4 \\
Quadruped & 2500000 & 0.0008 & 100 & 400 300 & 150 & 0.1 & 1000000 & 4 \\
Manipulation & 1200000 & 0.0003 & 100 & 128 96 & 45 & 0.25 & 250000 & 0.07
\end{tabular}
\caption{Hyperparameters for low level policy training. ``DM" stands for goal delta max, or the size of the goal space in each dimension sampled from during training.}
\label{tab:low_hparams}
\end{table*}

\begin{table*}[h]
\centering
\begin{tabular}{l|llllll}
Task & Timesteps & Learning Rate & Batch Size & Layers & Horizon & Buffer Size \\ \hline
Waypoint & 200000 & 0.0003 & 64 & 64 64 & 50 & 50000 \\
Maze & 400000 & 0.0003 & 64 & 64 64 & 100 & 50000 \\
Block Push & 500000 & 0.0003 & 64 & 64 64 & 60 & 50000 \\
Insert & 500000 & 0.0003 & 64 & 64 64 & 50 & 50000
\end{tabular}
\caption{Hyperparameters for high level policy training.}
\label{tab:high_hparams}
\end{table*}

\begin{table*}[h]
\centering
\begin{tabular}{l|llllll}
Agent A & Learning Rate & Batch Size & Layers & Update Freq & Weight & Stop \\ \hline
Nav PM & 0.0002 & 64 & 42 42 & 8 & 0.3 & 0.5 \\
PM, 2Link & 0.0003 & 64 & 42 42 & 8 & 0.4 & 0.5 \\
3Link & 0.0005 & 64 & 42 42 & 8 & 0.4 & 0.5
\end{tabular}
\caption{Hyperparameters for discriminator training.}
\label{tab:discirm_hparams}
\end{table*}

\section{Extended Zero-shot Results}
In our navigation experiments we also considered an additional agent, the \textit{Two-Leg} Ant. \textit{Waypoint} results for the \textit{Two-Legged Ant} can be found in Table \ref{tab:waypoint_zero_full} which contains complete zero-shot results with more precision. \textit{Maze} results can be found in Table \ref{tab:ant2_maze_zero}.

\begin{table*}[h]
\centering
\begin{tabular}{l|lllll}
 & Point Mass High & Ant High & Ant3 High & Ant2 High & Quadruped high \\ \hline
Point Mass Low & $1021.49 \pm 43.25$ & $602.56 \pm 33.82$ & $716.61 \pm 58.12$ & $593.18 \pm 60.09$ & $576.65 \pm 69.6$ \\
Ant Low & $482.72 \pm 38.96$ & $476.42 \pm 19.44$ & $472.85 \pm 50.68$ & $417.59 \pm 27.48$ & $406.96 \pm 35.63$ \\
Ant3 Low & $488.62 \pm 74.35$ & $483.59 \pm 71.67$ & $499.19 \pm 64.99$ & $471.24 \pm 65.42$ & $432.29 \pm 64.59$ \\
Ant2 Low & $353.56 \pm 39.33$ & $371.11 \pm 27.15$ & $388.38 \pm 31.56$ & $420.81 \pm 31.24$ & $373.99 \pm 34.52$ \\
Quadruped Low & $169.43 \pm 33.36$ & $182.33 \pm 21.55$ & $219.57 \pm 26.61$ & $267.12 \pm 18.95$ & $257.13 \pm 18.83$
\end{tabular}
\caption{Zero-Shot transfer for the way-point navigation task.}
\label{tab:waypoint_zero_full}
\end{table*}

Zero-shot results for the \textit{2-Link} arm were withheld from Table \ref{tab:zeroshot} for consistency with the \textit{PegInsert} task, which the two\textit{2-Link} arm was unable to complete due to its limited range of motion. Zero-shot results for the \textit{2-Link} on \textit{BlockPush} can be found in Table \ref{tab:2link_push_zero}.

\begin{table*}[h!]
\centering
\begin{tabular}{l|l|l|l|l}
Maze Task & PM High End   & Ant High End  & PM High Sample & Ant High Sample \\ \hline
Ant2 Low  & $.74 \pm .17$ & $.14 \pm .13$ & $.87 \pm .09$  & $.60 \pm .05$  
\end{tabular}
\caption{Zero-shot results for \textit{Two-Legged Ant} on \textit{Maze}}
\label{tab:ant2_maze_zero}
\end{table*}

\begin{table*}[h!]
\centering
\begin{tabular}{l|l|l|l|l}
Block Push 1 & PM HL   & 2Link HL  & 3Link HL & 4Link HL \\ \hline
Ant2 Low  & $.36 \pm .14$ & $.97 \pm .02$ & $.22 \pm .07$  & $.39 \pm .11$  
\end{tabular}
\caption{Zero-shot results for \textit{2-Link Arm} on \textit{Block Push 1}}
\label{tab:2link_push_zero}
\end{table*}

\section{Extended Discriminative Imitation Results}
In our initial experiments we considered both an online and offline data collection scheme used for training the discriminator. In the online version of data collection, roll-outs are collected from each agent running on the same goal $g$, ensuring the discriminator is trained on the same goals from both agents. Initial experiments showed that offline data collection, as described in section 4.2 was as good or better than online data collection in most cases. A possible explanation is that online data collection made the discriminator's task too easy. In the main body of the paper, we only report results from offline data collection. Here, Table \ref{tab:discrim_nav_full} and Table \ref{tab:discrim_push_full} contain results from all the online vs. offline comparisons we ran.

\section{Extended KL-regularized Finetuning Results}
\begin{figure*}[h!]
\centering
\includegraphics[width=0.6\linewidth]{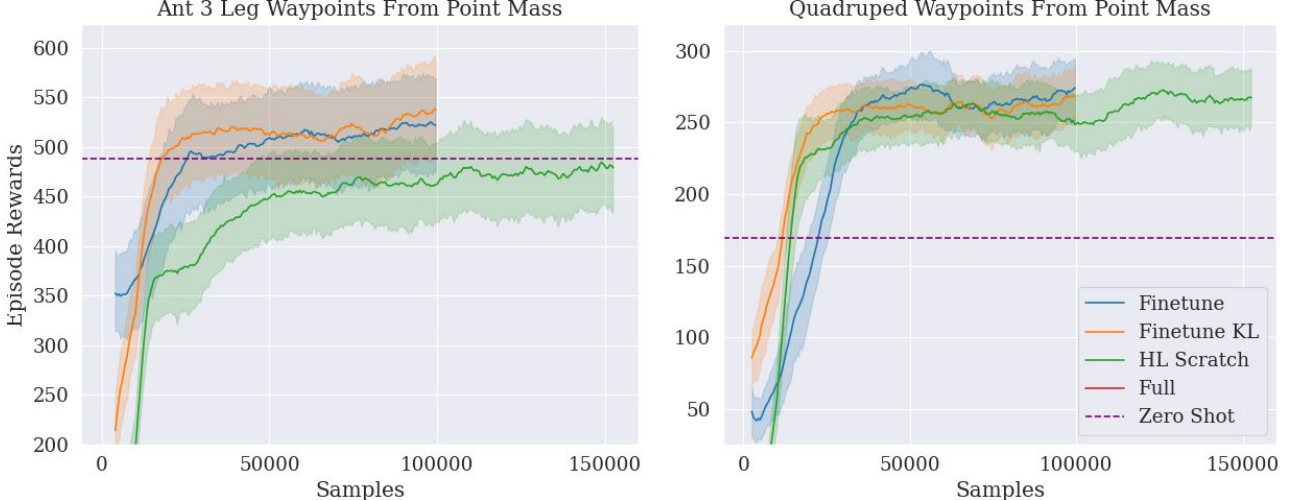}
\vspace{-0.15in}
\caption{Comparison of performance finetuning from \textit{PointMass} for \textit{Ant Waypoint},  and \textit{Quadruped Waypoint} respectively.}
\label{fig:extra_waypoint}
\end{figure*}

We ran finetuning experiments on the \textit{Waypoint} task that were not included in the main body of the paper due to their similarity to the included curve for the \textit{Ant} agent. Finetuning results for the \textit{3-Leg Ant} and the \textit{Quadruped} on the waypoint task are included in Figure \ref{fig:extra_waypoint}.

\begin{table*}[h!]
\centering
\begin{tabular}{l|l|l|l}
Task & Waypoint & Maze Sampled & Maze End \\ \hline
Transfer & PM$\vsarr$Ant & PM$\vsarr$Ant & PM$\vsarr$Ant \\ \hline
Zero-Shot & $482.72\pm38.96$ & $0.3\pm0.08$ & $0.65\pm0.04$ \\
Discrim Online & $467.22\pm20.61$ & $0.37\pm0.1$ & $0.63\pm0.04$ \\
Discrim Offline & $546.06\pm14.78$ & $0.55\pm0.13$ & $0.72\pm0.03$
\end{tabular}
\caption{Discriminative imitation zero-shot results for the Ant.}
\label{tab:discrim_nav_full}
\end{table*}

\begin{table*}[h!]
\resizebox{\textwidth}{!}{
\begin{tabular}{l|llll|llll}
Task & \multicolumn{4}{l|}{Block Push 1} & \multicolumn{4}{l}{Block Push 2} \\ \hline
Transfer & PM$\vsarr$3Link & 2Link$\vsarr$3Link & 2Link$\vsarr$4Link & 3Link$\vsarr$4Link & PM$\vsarr$3Link & 2Link$\vsarr$3Link & 2Link$\vsarr$4Link & 3Link$\vsarr$4Link \\ \hline
Zero-Shot & $0.17\pm.08$ & $0.2\pm.09$ & $0.07\pm.04$ & $0.1\pm.05$ & $0.24\pm.12$ & $0.61\pm.16$ & $0.19\pm.12$ & $0.39\pm.16$ \\
Discrim On & $0.34\pm.1$ & $0.43\pm.15$ & $0.17\pm.08$ & $0.23\pm.13$ & $0.43\pm.12$ & $0.42\pm.11$ & $0.46\pm.13$ & $0.44\pm.14$ \\
Discrim Off & $0.35\pm.13$ & $0.49\pm.12$ & $0.28\pm.14$ & $0.15\pm.04$ & $0.43\pm.15$ & $0.42\pm.11$ & $0.41\pm.16$ & $0.41\pm.15$ 
\end{tabular}
}
\caption{Discriminative imitation zero-shot results for various manipulation configurations.}
\label{tab:discrim_push_full}
\end{table*}

\section{Resources}
Our code can be found at \url{https://github.com/jhejna/hierarchical_morphology_transfer} and videos depicting results of our experiments can be found at \url{https://sites.google.com/berkeley.edu/morphology-transfer}.

%% file: main.bbl
\begin{thebibliography}{81}
\providecommand{\natexlab}[1]{#1}
\providecommand{\url}[1]{\texttt{#1}}
\expandafter\ifx\csname urlstyle\endcsname\relax
  \providecommand{\doi}[1]{doi: #1}\else
  \providecommand{\doi}{doi: \begingroup \urlstyle{rm}\Url}\fi

\bibitem[Abbeel \& Ng(2004)Abbeel and Ng]{abbeel2004apprenticeship}
Abbeel, P. and Ng, A.~Y.
\newblock Apprenticeship learning via inverse reinforcement learning.
\newblock In \emph{Proceedings of the twenty-first international conference on
  Machine learning}, pp.\ ~1, 2004.

\bibitem[Ammar et~al.(2015)Ammar, Eaton, Ruvolo, and
  Taylor]{ammar2015unsupervised}
Ammar, H.~B., Eaton, E., Ruvolo, P., and Taylor, M.~E.
\newblock Unsupervised cross-domain transfer in policy gradient reinforcement
  learning via manifold alignment.
\newblock In \emph{Twenty-Ninth AAAI Conference on Artificial Intelligence},
  2015.

\bibitem[Andreas et~al.(2016)Andreas, Klein, and Levine]{sketches}
Andreas, J., Klein, D., and Levine, S.
\newblock Modular multitask reinforcement learning with policy sketches, 2016.

\bibitem[Argall et~al.(2009)Argall, Chernova, Veloso, and
  Browning]{argall2009survey}
Argall, B.~D., Chernova, S., Veloso, M., and Browning, B.
\newblock A survey of robot learning from demonstration.
\newblock \emph{Robotics and autonomous systems}, 2009.

\bibitem[Aytar \& Zisserman(2011)Aytar and Zisserman]{aytar2011tabula}
Aytar, Y. and Zisserman, A.
\newblock Tabula rasa: Model transfer for object category detection.
\newblock In \emph{2011 international conference on computer vision}, pp.\
  2252--2259. IEEE, 2011.

\bibitem[Bacon et~al.(2017)Bacon, Harb, and Precup]{bacon2017option}
Bacon, P.-L., Harb, J., and Precup, D.
\newblock The option-critic architecture.
\newblock In \emph{Thirty-First AAAI Conference on Artificial Intelligence},
  2017.

\bibitem[Baker et~al.(2019)Baker, Kanitscheider, Markov, Wu, Powell, McGrew,
  and Mordatch]{baker2019emergent}
Baker, B., Kanitscheider, I., Markov, T., Wu, Y., Powell, G., McGrew, B., and
  Mordatch, I.
\newblock Emergent tool use from multi-agent autocurricula.
\newblock \emph{arXiv preprint arXiv:1909.07528}, 2019.

\bibitem[Barber \& Agakov(2004)Barber and Agakov]{barber2004information}
Barber, D. and Agakov, F.~V.
\newblock Information maximization in noisy channels: A variational approach.
\newblock In \emph{Advances in Neural Information Processing Systems}, pp.\
  201--208, 2004.

\bibitem[Barreto et~al.(2017)Barreto, Dabney, Munos, Hunt, Schaul, van Hasselt,
  and Silver]{barreto2017successor}
Barreto, A., Dabney, W., Munos, R., Hunt, J.~J., Schaul, T., van Hasselt,
  H.~P., and Silver, D.
\newblock Successor features for transfer in reinforcement learning.
\newblock In \emph{Advances in neural information processing systems}, pp.\
  4055--4065, 2017.

\bibitem[Barto \& Mahadevan(2003)Barto and Mahadevan]{barto2003recent}
Barto, A.~G. and Mahadevan, S.
\newblock Recent advances in hierarchical reinforcement learning.
\newblock \emph{Discrete event dynamic systems}, 13\penalty0 (1-2):\penalty0
  41--77, 2003.

\bibitem[Bengio et~al.(2009)Bengio, Louradour, Collobert, and
  Weston]{bengio2009curriculum}
Bengio, Y., Louradour, J., Collobert, R., and Weston, J.
\newblock Curriculum learning.
\newblock In \emph{ICML}, 2009.

\bibitem[Brockman et~al.(2016)Brockman, Cheung, Pettersson, Schneider,
  Schulman, Tang, and Zaremba]{brockman2016openai}
Brockman, G., Cheung, V., Pettersson, L., Schneider, J., Schulman, J., Tang,
  J., and Zaremba, W.
\newblock Openai gym.
\newblock \emph{arXiv preprint arXiv:1606.01540}, 2016.

\bibitem[Caruana(1997)]{caruana1997multitask}
Caruana, R.
\newblock Multitask learning.
\newblock \emph{Machine learning}, 28\penalty0 (1):\penalty0 41--75, 1997.

\bibitem[Chen et~al.(2018)Chen, Murali, and Gupta]{hcp}
Chen, T., Murali, A., and Gupta, A.
\newblock Hardware conditioned policies for multi-robot transfer learning.
\newblock In \emph{Advances in Neural Information Processing Systems}, pp.\
  9333--9344, 2018.

\bibitem[Devin et~al.(2017)Devin, Gupta, Darrell, Abbeel, and
  Levine]{devin2017learning}
Devin, C., Gupta, A., Darrell, T., Abbeel, P., and Levine, S.
\newblock Learning modular neural network policies for multi-task and
  multi-robot transfer.
\newblock In \emph{2017 IEEE International Conference on Robotics and
  Automation (ICRA)}, pp.\  2169--2176. IEEE, 2017.

\bibitem[Doersch \& Zisserman(2017)Doersch and Zisserman]{doersch2017multi}
Doersch, C. and Zisserman, A.
\newblock Multi-task self-supervised visual learning.
\newblock In \emph{Proceedings of the IEEE International Conference on Computer
  Vision}, pp.\  2051--2060, 2017.

\bibitem[Duan et~al.(2016)Duan, Chen, Houthooft, Schulman, and
  Abbeel]{duan2016benchmarking}
Duan, Y., Chen, X., Houthooft, R., Schulman, J., and Abbeel, P.
\newblock Benchmarking deep reinforcement learning for continuous control.
\newblock In \emph{International Conference on Machine Learning}, pp.\
  1329--1338, 2016.

\bibitem[Esmaili et~al.(1995)Esmaili, Sammut, and
  Shirazi]{esmaili1995behavioural}
Esmaili, N., Sammut, C., and Shirazi, G.
\newblock Behavioural cloning in control of a dynamic system.
\newblock IEEE, 1995.

\bibitem[Eysenbach et~al.(2018)Eysenbach, Gupta, Ibarz, and Levine]{diversity}
Eysenbach, B., Gupta, A., Ibarz, J., and Levine, S.
\newblock Diversity is all you need: Learning skills without a reward function.
\newblock \emph{CoRR}, abs/1802.06070, 2018.
\newblock URL \url{http://arxiv.org/abs/1802.06070}.

\bibitem[Fernando et~al.(2013)Fernando, Habrard, Sebban, and
  Tuytelaars]{fernando2013unsupervised}
Fernando, B., Habrard, A., Sebban, M., and Tuytelaars, T.
\newblock Unsupervised visual domain adaptation using subspace alignment.
\newblock In \emph{Proceedings of the IEEE international conference on computer
  vision}, pp.\  2960--2967, 2013.

\bibitem[Florensa et~al.(2017)Florensa, Duan, and Abbeel]{snn}
Florensa, C., Duan, Y., and Abbeel, P.
\newblock Stochastic neural networks for hierarchical reinforcement learning,
  2017.

\bibitem[Galashov et~al.(2019)Galashov, Jayakumar, Hasenclever, Tirumala,
  Schwarz, Desjardins, Czarnecki, Teh, Pascanu, and
  Heess]{galashov2019information}
Galashov, A., Jayakumar, S.~M., Hasenclever, L., Tirumala, D., Schwarz, J.,
  Desjardins, G., Czarnecki, W.~M., Teh, Y.~W., Pascanu, R., and Heess, N.
\newblock Information asymmetry in kl-regularized rl.
\newblock \emph{arXiv preprint arXiv:1905.01240}, 2019.

\bibitem[Girshick et~al.(2014)Girshick, Donahue, Darrell, and
  Malik]{girshick2014rich}
Girshick, R., Donahue, J., Darrell, T., and Malik, J.
\newblock Rich feature hierarchies for accurate object detection and semantic
  segmentation.
\newblock In \emph{Proceedings of the IEEE conference on computer vision and
  pattern recognition}, pp.\  580--587, 2014.

\bibitem[Gopalan et~al.(2011)Gopalan, Li, and Chellappa]{gopalan2011domain}
Gopalan, R., Li, R., and Chellappa, R.
\newblock Domain adaptation for object recognition: An unsupervised approach.
\newblock In \emph{2011 international conference on computer vision}, pp.\
  999--1006. IEEE, 2011.

\bibitem[Goyal et~al.(2019)Goyal, Islam, Strouse, Ahmed, Botvinick, Larochelle,
  Bengio, and Levine]{goyal2019infobot}
Goyal, A., Islam, R., Strouse, D., Ahmed, Z., Botvinick, M., Larochelle, H.,
  Bengio, Y., and Levine, S.
\newblock Infobot: Transfer and exploration via the information bottleneck.
\newblock \emph{arXiv preprint arXiv:1901.10902}, 2019.

\bibitem[Haarnoja et~al.(2018)Haarnoja, Zhou, Hartikainen, Tucker, Ha, Tan,
  Kumar, Zhu, Gupta, Abbeel, et~al.]{haarnoja2018soft}
Haarnoja, T., Zhou, A., Hartikainen, K., Tucker, G., Ha, S., Tan, J., Kumar,
  V., Zhu, H., Gupta, A., Abbeel, P., et~al.
\newblock Soft actor-critic algorithms and applications.
\newblock \emph{arXiv preprint arXiv:1812.05905}, 2018.

\bibitem[Hill et~al.(2018)Hill, Raffin, Ernestus, Gleave, Kanervisto, Traore,
  Dhariwal, Hesse, Klimov, Nichol, Plappert, Radford, Schulman, Sidor, and
  Wu]{stable-baselines}
Hill, A., Raffin, A., Ernestus, M., Gleave, A., Kanervisto, A., Traore, R.,
  Dhariwal, P., Hesse, C., Klimov, O., Nichol, A., Plappert, M., Radford, A.,
  Schulman, J., Sidor, S., and Wu, Y.
\newblock Stable baselines.
\newblock \url{https://github.com/hill-a/stable-baselines}, 2018.

\bibitem[Ho \& Ermon(2016)Ho and Ermon]{ho2016generative}
Ho, J. and Ermon, S.
\newblock Generative adversarial imitation learning.
\newblock In \emph{Advances in neural information processing systems}, pp.\
  4565--4573, 2016.

\bibitem[Hoffman et~al.(2014)Hoffman, Darrell, and
  Saenko]{hoffman2014continuous}
Hoffman, J., Darrell, T., and Saenko, K.
\newblock Continuous manifold based adaptation for evolving visual domains.
\newblock In \emph{Proceedings of the IEEE Conference on Computer Vision and
  Pattern Recognition}, pp.\  867--874, 2014.

\bibitem[Hu \& Montana(2019)Hu and Montana]{hu2019skill}
Hu, Y. and Montana, G.
\newblock Skill transfer in deep reinforcement learning under morphological
  heterogeneity.
\newblock \emph{arXiv preprint arXiv:1908.05265}, 2019.

\bibitem[Jaderberg et~al.(2016)Jaderberg, Mnih, Czarnecki, Schaul, Leibo,
  Silver, and Kavukcuoglu]{jaderberg2016reinforcement}
Jaderberg, M., Mnih, V., Czarnecki, W.~M., Schaul, T., Leibo, J.~Z., Silver,
  D., and Kavukcuoglu, K.
\newblock Reinforcement learning with unsupervised auxiliary tasks.
\newblock \emph{arXiv preprint arXiv:1611.05397}, 2016.

\bibitem[Jhuo et~al.(2012)Jhuo, Liu, Lee, and Chang]{jhuo2012robust}
Jhuo, I.-H., Liu, D., Lee, D., and Chang, S.-F.
\newblock Robust visual domain adaptation with low-rank reconstruction.
\newblock In \emph{2012 IEEE conference on computer vision and pattern
  recognition}, pp.\  2168--2175. IEEE, 2012.

\bibitem[Kaelbling et~al.(1996)Kaelbling, Littman, and
  Moore]{kaelbling1996reinforcement}
Kaelbling, L.~P., Littman, M.~L., and Moore, A.~W.
\newblock Reinforcement learning: A survey.
\newblock \emph{Journal of artificial intelligence research}, 4:\penalty0
  237--285, 1996.

\bibitem[Kakade(2002)]{kakade2002natural}
Kakade, S.~M.
\newblock A natural policy gradient.
\newblock In \emph{Advances in neural information processing systems}, pp.\
  1531--1538, 2002.

\bibitem[Kansky et~al.(2017)Kansky, Silver, M{\'e}ly, Eldawy,
  L{\'a}zaro-Gredilla, Lou, Dorfman, Sidor, Phoenix, and
  George]{kansky2017schema}
Kansky, K., Silver, T., M{\'e}ly, D.~A., Eldawy, M., L{\'a}zaro-Gredilla, M.,
  Lou, X., Dorfman, N., Sidor, S., Phoenix, S., and George, D.
\newblock Schema networks: Zero-shot transfer with a generative causal model of
  intuitive physics.
\newblock In \emph{Proceedings of the 34th International Conference on Machine
  Learning-Volume 70}, pp.\  1809--1818. JMLR. org, 2017.

\bibitem[Kober \& Peters(2009)Kober and Peters]{kober2009learning}
Kober, J. and Peters, J.
\newblock Learning motor primitives for robotics.
\newblock In \emph{ICRA}, 2009.

\bibitem[Kokkinos(2017)]{kokkinos2017ubernet}
Kokkinos, I.
\newblock Ubernet: Training a universal convolutional neural network for low-,
  mid-, and high-level vision using diverse datasets and limited memory.
\newblock In \emph{Proceedings of the IEEE Conference on Computer Vision and
  Pattern Recognition}, pp.\  6129--6138, 2017.

\bibitem[Krishnan et~al.(2018)Krishnan, Garg, Patil, Lea, Hager, Abbeel, and
  Goldberg]{krishnan2018transition}
Krishnan, S., Garg, A., Patil, S., Lea, C., Hager, G., Abbeel, P., and
  Goldberg, K.
\newblock Transition state clustering: Unsupervised surgical trajectory
  segmentation for robot learning.
\newblock In \emph{RR}. 2018.

\bibitem[Kulis et~al.(2011)Kulis, Saenko, and Darrell]{kulis2011you}
Kulis, B., Saenko, K., and Darrell, T.
\newblock What you saw is not what you get: Domain adaptation using asymmetric
  kernel transforms.
\newblock In \emph{CVPR 2011}, pp.\  1785--1792. IEEE, 2011.

\bibitem[Kulkarni et~al.(2016)Kulkarni, Narasimhan, Saeedi, and
  Tenenbaum]{kulkarni2016hierarchical}
Kulkarni, T.~D., Narasimhan, K., Saeedi, A., and Tenenbaum, J.
\newblock Hierarchical deep reinforcement learning: Integrating temporal
  abstraction and intrinsic motivation.
\newblock In \emph{Advances in neural information processing systems}, pp.\
  3675--3683, 2016.

\bibitem[Levy et~al.(2017)Levy, Konidaris, Platt, and Saenko]{hac}
Levy, A., Konidaris, G., Platt, R., and Saenko, K.
\newblock Learning multi-level hierarchies with hindsight, 2017.

\bibitem[Li et~al.(2019{\natexlab{a}})Li, Florensa, Clavera, and
  Abbeel]{li2019sub}
Li, A.~C., Florensa, C., Clavera, I., and Abbeel, P.
\newblock Sub-policy adaptation for hierarchical reinforcement learning.
\newblock \emph{arXiv preprint arXiv:1906.05862}, 2019{\natexlab{a}}.

\bibitem[Li et~al.(2019{\natexlab{b}})Li, Wang, Tang, and
  Zhang]{li2019hierarchical}
Li, S., Wang, R., Tang, M., and Zhang, C.
\newblock Hierarchical reinforcement learning with advantage-based auxiliary
  rewards.
\newblock In \emph{Advances in Neural Information Processing Systems}, pp.\
  1407--1417, 2019{\natexlab{b}}.

\bibitem[Li et~al.(2017)Li, Song, and Ermon]{li2017infogail}
Li, Y., Song, J., and Ermon, S.
\newblock Infogail: Interpretable imitation learning from visual
  demonstrations.
\newblock In \emph{Advances in Neural Information Processing Systems}, pp.\
  3812--3822, 2017.

\bibitem[Lillicrap et~al.(2015)Lillicrap, Hunt, Pritzel, Heess, Erez, Tassa,
  Silver, and Wierstra]{lillicrap2015continuous}
Lillicrap, T.~P., Hunt, J.~J., Pritzel, A., Heess, N., Erez, T., Tassa, Y.,
  Silver, D., and Wierstra, D.
\newblock Continuous control with deep reinforcement learning.
\newblock \emph{arXiv preprint arXiv:1509.02971}, 2015.

\bibitem[Marino et~al.(2018)Marino, Gupta, Fergus, and
  Szlam]{marino2018hierarchical}
Marino, K., Gupta, A., Fergus, R., and Szlam, A.
\newblock Hierarchical rl using an ensemble of proprioceptive periodic
  policies.
\newblock 2018.

\bibitem[Meier et~al.(2011)Meier, Theodorou, Stulp, and
  Schaal]{meier2011movement}
Meier, F., Theodorou, E., Stulp, F., and Schaal, S.
\newblock Movement segmentation using a primitive library.
\newblock In \emph{2011 IEEE/RSJ International Conference on Intelligent Robots
  and Systems}, pp.\  3407--3412. IEEE, 2011.

\bibitem[Murali et~al.(2016)Murali, Garg, Krishnan, Pokorny, Abbeel, Darrell,
  and Goldberg]{murali2016tsc}
Murali, A., Garg, A., Krishnan, S., Pokorny, F.~T., Abbeel, P., Darrell, T.,
  and Goldberg, K.
\newblock Tsc-dl: Unsupervised trajectory segmentation of multi-modal surgical
  demonstrations with deep learning.
\newblock In \emph{ICRA}, 2016.

\bibitem[Murali et~al.(2018)Murali, Pinto, Gandhi, and Gupta]{murali2018cassl}
Murali, A., Pinto, L., Gandhi, D., and Gupta, A.
\newblock Cassl: Curriculum accelerated self-supervised learning.
\newblock In \emph{2018 IEEE International Conference on Robotics and
  Automation (ICRA)}, pp.\  6453--6460. IEEE, 2018.

\bibitem[Nachum et~al.(2018{\natexlab{a}})Nachum, Gu, Lee, and Levine]{hiro}
Nachum, O., Gu, S., Lee, H., and Levine, S.
\newblock Data-efficient hierarchical reinforcement learning.
\newblock \emph{CoRR}, abs/1805.08296, 2018{\natexlab{a}}.
\newblock URL \url{http://arxiv.org/abs/1805.08296}.

\bibitem[Nachum et~al.(2018{\natexlab{b}})Nachum, Gu, Lee, and
  Levine]{nachum2018near}
Nachum, O., Gu, S., Lee, H., and Levine, S.
\newblock Near-optimal representation learning for hierarchical reinforcement
  learning.
\newblock \emph{arXiv preprint arXiv:1810.01257}, 2018{\natexlab{b}}.

\bibitem[Ng et~al.(2000)Ng, Russell, et~al.]{ng2000algorithms}
Ng, A.~Y., Russell, S.~J., et~al.
\newblock Algorithms for inverse reinforcement learning.
\newblock In \emph{Icml}, volume~1, pp.\  663--670, 2000.

\bibitem[Nicolescu \& Mataric(2003)Nicolescu and Mataric]{nicolescu2003natural}
Nicolescu, M.~N. and Mataric, M.~J.
\newblock Natural methods for robot task learning: Instructive demonstrations,
  generalization and practice.
\newblock In \emph{AAMAS}, 2003.

\bibitem[Niekum et~al.(2012)Niekum, Osentoski, Konidaris, and
  Barto]{niekum2012learning}
Niekum, S., Osentoski, S., Konidaris, G., and Barto, A.~G.
\newblock Learning and generalization of complex tasks from unstructured
  demonstrations.
\newblock IEEE, 2012.

\bibitem[Omidshafiei et~al.(2017)Omidshafiei, Pazis, Amato, How, and
  Vian]{omidshafiei2017deep}
Omidshafiei, S., Pazis, J., Amato, C., How, J.~P., and Vian, J.
\newblock Deep decentralized multi-task multi-agent reinforcement learning
  under partial observability.
\newblock In \emph{Proceedings of the 34th International Conference on Machine
  Learning-Volume 70}, pp.\  2681--2690. JMLR. org, 2017.

\bibitem[Pan \& Yang(2009)Pan and Yang]{pan2009survey}
Pan, S.~J. and Yang, Q.
\newblock A survey on transfer learning.
\newblock \emph{IEEE Transactions on knowledge and data engineering},
  22\penalty0 (10):\penalty0 1345--1359, 2009.

\bibitem[Parisotto et~al.(2015)Parisotto, Ba, and
  Salakhutdinov]{parisotto2015actor}
Parisotto, E., Ba, J.~L., and Salakhutdinov, R.
\newblock Actor-mimic: Deep multitask and transfer reinforcement learning.
\newblock \emph{arXiv preprint arXiv:1511.06342}, 2015.

\bibitem[Pathak et~al.(2019)Pathak, Lu, Darrell, Isola, and
  Efros]{pathak2019learning}
Pathak, D., Lu, C., Darrell, T., Isola, P., and Efros, A.~A.
\newblock Learning to control self-assembling morphologies: a study of
  generalization via modularity.
\newblock In \emph{Advances in Neural Information Processing Systems}, pp.\
  2292--2302, 2019.

\bibitem[Peng et~al.(2017)Peng, Berseth, Yin, and Van
  De~Panne]{peng2017deeploco}
Peng, X.~B., Berseth, G., Yin, K., and Van De~Panne, M.
\newblock Deeploco: Dynamic locomotion skills using hierarchical deep
  reinforcement learning.
\newblock \emph{ACM Transactions on Graphics (TOG)}, 36\penalty0 (4):\penalty0
  1--13, 2017.

\bibitem[Peng et~al.(2019)Peng, Chang, Zhang, Abbeel, and Levine]{peng2019mcp}
Peng, X.~B., Chang, M., Zhang, G., Abbeel, P., and Levine, S.
\newblock Mcp: Learning composable hierarchical control with multiplicative
  compositional policies.
\newblock In \emph{Advances in Neural Information Processing Systems}, pp.\
  3681--3692, 2019.

\bibitem[Peters et~al.(2013)Peters, Kober, M{\"u}lling, Kr{\"a}mer, and
  Neumann]{peters2013towards}
Peters, J., Kober, J., M{\"u}lling, K., Kr{\"a}mer, O., and Neumann, G.
\newblock Towards robot skill learning: From simple skills to table tennis.
\newblock In \emph{Joint European Conference on Machine Learning and Knowledge
  Discovery in Databases}, pp.\  627--631. Springer, 2013.

\bibitem[Pinto \& Gupta(2017)Pinto and Gupta]{pinto2017learning}
Pinto, L. and Gupta, A.
\newblock Learning to push by grasping: Using multiple tasks for effective
  learning.
\newblock In \emph{2017 IEEE International Conference on Robotics and
  Automation (ICRA)}, pp.\  2161--2168. IEEE, 2017.

\bibitem[Rajeswaran et~al.(2017)Rajeswaran, Kumar, Gupta, Vezzani, Schulman,
  Todorov, and Levine]{rajeswaran2017learning}
Rajeswaran, A., Kumar, V., Gupta, A., Vezzani, G., Schulman, J., Todorov, E.,
  and Levine, S.
\newblock Learning complex dexterous manipulation with deep reinforcement
  learning and demonstrations, 2017.

\bibitem[Rusu et~al.(2016)Rusu, Rabinowitz, Desjardins, Soyer, Kirkpatrick,
  Kavukcuoglu, Pascanu, and Hadsell]{rusu2016progressive}
Rusu, A.~A., Rabinowitz, N.~C., Desjardins, G., Soyer, H., Kirkpatrick, J.,
  Kavukcuoglu, K., Pascanu, R., and Hadsell, R.
\newblock Progressive neural networks.
\newblock \emph{arXiv preprint arXiv:1606.04671}, 2016.

\bibitem[Saenko et~al.(2010)Saenko, Kulis, Fritz, and
  Darrell]{saenko2010adapting}
Saenko, K., Kulis, B., Fritz, M., and Darrell, T.
\newblock Adapting visual category models to new domains.
\newblock In \emph{European conference on computer vision}, pp.\  213--226.
  Springer, 2010.

\bibitem[Schulman et~al.(2015)Schulman, Levine, Abbeel, Jordan, and
  Moritz]{schulman2015trust}
Schulman, J., Levine, S., Abbeel, P., Jordan, M.~I., and Moritz, P.
\newblock Trust region policy optimization.
\newblock In \emph{ICML}, pp.\  1889--1897, 2015.

\bibitem[Schulman et~al.(2017)Schulman, Wolski, Dhariwal, Radford, and
  Klimov]{schulman2017proximal}
Schulman, J., Wolski, F., Dhariwal, P., Radford, A., and Klimov, O.
\newblock Proximal policy optimization algorithms.
\newblock \emph{arXiv preprint arXiv:1707.06347}, 2017.

\bibitem[Shankar et~al.(2020)Shankar, Tulsiani, Pinto, and
  Gupta]{shankar2020discovering}
Shankar, T., Tulsiani, S., Pinto, L., and Gupta, A.
\newblock Discovering motor programs by recomposing demonstrations.
\newblock In \emph{International Conference on Learning Representations}, 2020.
\newblock URL \url{https://openreview.net/forum?id=rkgHY0NYwr}.

\bibitem[Sharma et~al.(2019{\natexlab{a}})Sharma, Gu, Levine, Kumar, and
  Hausman]{DADS}
Sharma, A., Gu, S., Levine, S., Kumar, V., and Hausman, K.
\newblock Dynamics-aware unsupervised discovery of skills.
\newblock \emph{CoRR}, abs/1907.01657, 2019{\natexlab{a}}.
\newblock URL \url{http://arxiv.org/abs/1907.01657}.

\bibitem[Sharma et~al.(2019{\natexlab{b}})Sharma, Pathak, and
  Gupta]{sharma2019third}
Sharma, P., Pathak, D., and Gupta, A.
\newblock Third-person visual imitation learning via decoupled hierarchical
  controller.
\newblock In \emph{Advances in Neural Information Processing Systems}, pp.\
  2593--2603, 2019{\natexlab{b}}.

\bibitem[Stadie et~al.(2017)Stadie, Abbeel, and Sutskever]{stadie2017third}
Stadie, B.~C., Abbeel, P., and Sutskever, I.
\newblock Third-person imitation learning.
\newblock \emph{arXiv preprint arXiv:1703.01703}, 2017.

\bibitem[Sutton et~al.(1998)Sutton, Barto, et~al.]{sutton1998introduction}
Sutton, R.~S., Barto, A.~G., et~al.
\newblock \emph{Introduction to reinforcement learning}, volume~2.
\newblock MIT press Cambridge, 1998.

\bibitem[Tassa et~al.(2018)Tassa, Doron, Muldal, Erez, Li, Casas, Budden,
  Abdolmaleki, Merel, Lefrancq, et~al.]{tassa2018deepmind}
Tassa, Y., Doron, Y., Muldal, A., Erez, T., Li, Y., Casas, D. d.~L., Budden,
  D., Abdolmaleki, A., Merel, J., Lefrancq, A., et~al.
\newblock Deepmind control suite.
\newblock \emph{arXiv preprint arXiv:1801.00690}, 2018.

\bibitem[Taylor \& Stone(2009)Taylor and Stone]{taylor2009transfer}
Taylor, M.~E. and Stone, P.
\newblock Transfer learning for reinforcement learning domains: A survey.
\newblock \emph{Journal of Machine Learning Research}, 10\penalty0
  (Jul):\penalty0 1633--1685, 2009.

\bibitem[Tirumala et~al.(2019)Tirumala, Noh, Galashov, Hasenclever, Ahuja,
  Wayne, Pascanu, Teh, and Heess]{tirumala2019exploiting}
Tirumala, D., Noh, H., Galashov, A., Hasenclever, L., Ahuja, A., Wayne, G.,
  Pascanu, R., Teh, Y.~W., and Heess, N.
\newblock Exploiting hierarchy for learning and transfer in kl-regularized rl.
\newblock \emph{arXiv preprint arXiv:1903.07438}, 2019.

\bibitem[Todorov et~al.(2012)Todorov, Erez, and Tassa]{todorov2012mujoco}
Todorov, E., Erez, T., and Tassa, Y.
\newblock Mujoco: A physics engine for model-based control.
\newblock In \emph{2012 IEEE/RSJ International Conference on Intelligent Robots
  and Systems}, pp.\  5026--5033. IEEE, 2012.

\bibitem[Torrey \& Shavlik(2010)Torrey and Shavlik]{torrey2010transfer}
Torrey, L. and Shavlik, J.
\newblock Transfer learning.
\newblock In \emph{Handbook of research on machine learning applications and
  trends: algorithms, methods, and techniques}, pp.\  242--264. IGI Global,
  2010.

\bibitem[Tzeng et~al.(2017)Tzeng, Hoffman, Saenko, and
  Darrell]{tzeng2017adversarial}
Tzeng, E., Hoffman, J., Saenko, K., and Darrell, T.
\newblock Adversarial discriminative domain adaptation.
\newblock In \emph{Proceedings of the IEEE Conference on Computer Vision and
  Pattern Recognition}, pp.\  7167--7176, 2017.

\bibitem[Williams(1992)]{williams1992simple}
Williams, R.~J.
\newblock Simple statistical gradient-following algorithms for connectionist
  reinforcement learning.
\newblock \emph{Machine learning}, 8\penalty0 (3-4):\penalty0 229--256, 1992.

\bibitem[Wolpert \& Kawato(1998)Wolpert and Kawato]{wolpert1998multiple}
Wolpert, D.~M. and Kawato, M.
\newblock Multiple paired forward and inverse models for motor control.
\newblock \emph{Neural networks}, 11\penalty0 (7-8):\penalty0 1317--1329, 1998.

\bibitem[Yang et~al.(2007)Yang, Yan, and Hauptmann]{yang2007adapting}
Yang, J., Yan, R., and Hauptmann, A.~G.
\newblock Adapting svm classifiers to data with shifted distributions.
\newblock In \emph{Seventh IEEE International Conference on Data Mining
  Workshops (ICDMW 2007)}, pp.\  69--76. IEEE, 2007.

\end{thebibliography}
